\documentclass[runningheads]{llncs}

\usepackage{eccv}

\usepackage{eccvabbrv}

\usepackage[accsupp]{axessibility}  
\usepackage{graphicx}
\usepackage{amsmath}
\usepackage{amssymb}
\usepackage{booktabs} 
\usepackage[normalem]{ulem}
\useunder{\uline}{\ul}{} 
\usepackage[table]{xcolor}
\usepackage{colortbl} 
\usepackage{multirow} 
\usepackage{booktabs}
\usepackage{caption} 
\usepackage{subcaption}

\usepackage{etoolbox}
\usepackage{silence}
\makeatletter
\robustify\@latex@warning@no@line
\makeatother
\usepackage{authblk}

\renewcommand*{\Affilfont}{\normalsize}

\makeatletter
\renewcommand\maketitle{\AB@maketitle} 
\renewcommand\AB@affilsepx{\quad\protect\Affilfont} 
\makeatother 
\usepackage{orcidlink}
\usepackage{fancyhdr}

\usepackage{hyperref}

\begin{document}

\title{UniH$^3$: Unifying Hierarchical Homogeneity and Heterogeneity for All-in-One Medical Image Restoration} 

\titlerunning{Abbreviated paper title}


\author{Zhiwen Yang\inst{1}\orcidlink{0000-0002-5712-9695} \and
Jiayin Li\inst{1}\orcidlink{0009-0006-9021-4230} \and Chengyu Liu\inst{1}\orcidlink{0009-0002-6045-3346
} \and
Hui Zhang\inst{2}\orcidlink{0000-0001-9476-4244} \and Bingzheng Wei\inst{3}\orcidlink{0000-0001-6979-0459} \and Yan Xu\inst{1}$^\dag$\orcidlink{0000-0002-2636-7594}}

\authorrunning{Yang et al.}

\institute{School of Biological Science and Medical Engineering, Beihang University, Beijing 100191, China \\ \and
Department of Biomedical Engineering, Tsinghua University, Beijing 100084, China \\
 \and
Independent Researcher\\  
$^\dag$ Corresponding Author \\  
\email{upyzwup@buaa.edu.cn xuyan04@gmail.com}}

\maketitle

\begin{abstract}
All-in-One medical image restoration (MedIR) aims to address diverse tasks across modalities and degradation types using a single universal model. Existing methods typically prioritize modeling inter-task heterogeneity (e.g., distinct data distributions and degradation types). However, they largely neglect the inherent homogeneity present in medical images, such as widely shared anatomical structures within and across modalities, which can be leveraged to ease model training and improve generalization. To this end, we propose \textbf{UniH$^3$}, a novel framework that \textbf{Uni}fies \textbf{H}ierarchical \textbf{H}omogeneity and \textbf{H}eterogeneity for all-in-one medical image restoration. Specifically, to comprehensively exploit homogeneity, we introduce a Hierarchical Homogeneity Memory (H$^2$M) module that progressively distills intra- and inter-task homogeneity priors from high-quality images during training, and adaptively retrieves the most relevant priors tailored to the input for guided restoration. These retrieved priors are then injected into the restoration pipeline via an efficient Homogeneity-Guided Attention (HGA) mechanism. Furthermore, to comprehensively address heterogeneity, we design a Hierarchical Heterogeneity Balancer (H$^2$B) that mitigates both inter- and intra-task conflicts during optimization, facilitating balanced and effective multi-task learning. Extensive experiments on two large-scale benchmarks—MedIR-2D-500K and MedIR-3D-3K—demonstrate that UniH$^3$ achieves state-of-the-art performance on both all-in-one and single-task medical image restoration. We hope this work establishes a strong benchmark and advances the development of general-purpose medical image restoration models. Code is available at \href{https://github.com/Yaziwel/UniH3.git}{https://github.com/Yaziwel/UniH3}.

\keywords{Medical Image Restoration \and All-in-One \and Universal Model}
\end{abstract}

\section{Introduction}
\label{sec_intro}
Medical image restoration (MedIR) aims to recover a high-quality (HQ) image from a degraded low-quality (LQ) acquisition. Since each medical image modality (e.g., PET, CT, MRI) operates under distinct physical principles and is often studied independently, most MedIR research has focused on a single-task setting, in which researchers train specialized models to address the primary degradation introduced by the imaging physics of each modality. Typical MedIR includes PET image denoising \cite{wang20183dcgan,zhou2022sgsgan,yang2023drmc,yang2026unipet}, CT image denoising \cite{chen2017redcnn,wang2023ctformer,ozturk2024denomamba}, MRI image super-resolution \cite{chen2018mdcsrn,wang2020mrdg,huang2022swinmr,song2025unicoal}. Despite their success in specific scenarios, single-task models have limited practicality for two main reasons. First, in complex scenarios where multiple MedIR tasks coexist (e.g., multimodal PET/CT and PET/MRI), single-task models trained for one task often underperform on others. Moreover, training separate models for each task leads to inefficiencies in both deployment and maintenance. Second, the single-task paradigm hinders progress toward more general intelligence in MedIR. These limitations motivate interest in a universal model that can handle diverse MedIR tasks.

Recent advances in computer vision have fostered the emergence of All-in-One restoration frameworks~\cite{valanarasu2022transweather,li2022airnet,yang2024amir,potlapalli2023promptir,cui2025adair,yang2025tat,chen2025endoir}. Pioneering research in the medical domain~\cite{yang2024amir,yang2025tat,chen2025diffcode,chen2025endoir}, particularly the first work on AMIR~\cite{yang2024amir}, has established the feasibility of unified modeling for MedIR. To manage diverse tasks within a single model, existing approaches predominantly focus on modeling task heterogeneity—that is, distinguishing between tasks to apply specialized processing. Techniques such as contrastive learning~\cite{li2022airnet}, degradation classification~\cite{hu2025degradation_classification}, visual prompting~\cite{potlapalli2023promptir}, and mixture-of-experts (MoE)~\cite{yang2024amir,zamfir2025moceir} are widely employed to distinguish between tasks.

However, we argue that current All-in-One methods suffer from two critical limitations. 
\textbf{First}, they largely overlook the inherent homogeneity of medical images. Compared with natural images, medical images from different modalities and tasks often exhibit more consistent anatomical structures and share stronger biological priors. Neglecting this shared knowledge prevents models from exploiting cross-task synergies, thereby increasing the difficulty of learning as the number of tasks grows.
\textbf{Second}, regarding heterogeneity, existing methods typically address only inter-task differences (e.g., different degradation types or modalities) while ignoring intra-task variations (e.g., variations caused by different scanners, centers, or patient demographics). This coarse-grained approach fails to resolve optimization conflicts that arise from subtle intra-task distribution shifts. Therefore, it is imperative to simultaneously model both homogeneity and heterogeneity at a hierarchical level (inter- and intra-task) to achieve robust and effective All-in-One MedIR.

To address these challenges, we propose \textbf{UniH$^3$}, a novel framework that \textbf{Uni}fies \textbf{H}ierarchical \textbf{H}omogeneity and \textbf{H}eterogeneity for All-in-One medical image restoration. 
On the one hand, to fully exploit shared knowledge, we introduce a Hierarchical Homogeneity Memory (H$^2$M) module. This module progressively distills both inter- and intra-task priors from HQ images into a memory bank during training. During inference, it adaptively retrieves the most relevant structural priors tailored to the input, which are then injected into the network via an efficient Homogeneity-Guided Attention (HGA) mechanism to guide restoration.
On the other hand, to manage task conflicts comprehensively, we design a Hierarchical Heterogeneity Balancer (H$^2$B). Unlike traditional weighting strategies~\cite{kendall2018uncertainty_loss,wu2025debiased_uncertainty_loss} that only balance loss functions at the task level, H$^2$B dynamically mitigates optimization conflicts at both inter- and intra-task levels, ensuring balanced convergence across diverse data distributions. Finally, to validate the effectiveness of UniH$^3$, we construct a benchmark comprising two large-scale datasets: MedIR-2D-500K, containing 509,200 2D image pairs across seven 2D MedIR tasks, and MedIR-3D-3K, containing 3,522 3D volume pairs across three 3D MedIR tasks. Extensive experiments on this benchmark indicate that UniH$^3$ achieves state-of-the-art (SOTA) performance in both all-in-one and single-task medical image restoration. 

In summary, our contributions are as follows:
\begin{itemize}
\item  We propose UniH$^3$, a novel framework that simultaneously models hierarchical inter- and intra-task homogeneity and heterogeneity for effective all-in-one medical image restoration.
\item  We present a Hierarchical Homogeneity Memory (H$^2$M) module that can adaptively distill and retrieve homogeneity priors to guide the restoration process. Additionally, an efficient Homogeneity-Guided Attention (HGA) mechanism is introduced to fully exploit the retrieved prior for guided restoration.
\item  We develop a Hierarchical Heterogeneity Balancer (H$^2$B), which achieves fine-grained task balancing by resolving optimization conflicts arising from both inter-task distinctions and intra-task variations.
\end{itemize}

\section{Related Work} 

\textbf{Single-Task Medical Image Restoration.} Because different medical imaging modalities are typically studied independently, most MedIR research focuses on single-task problems that address the primary degradations encountered in each modality. Typical MedIR tasks include PET image denoising \cite{wang20183dcgan,zhou2022sgsgan,yang2023drmc,yang2026unipet}, CT denoising \cite{chen2017redcnn,wang2023ctformer,ozturk2024denomamba}, MRI super-resolution \cite{chen2018mdcsrn,wang2020mrdg,huang2022swinmr}, X-ray denoising \cite{thanh2019x_ray_denoising}, OCT denoising \cite{dong2020oct_denoising}, ultrasound denoising \cite{asgariandehkordi2023ultrsound}, and pathology image super-resolution \cite{li2021pathsr}. With the development of deep learning—especially recent advances in network architectures such as convolutional neural networks (CNNs) \cite{lecun1989cnn,chen2017redcnn}, Transformers \cite{vaswani2017transformer,yang2024rat}, Mamba \cite{gu2023mamba,ozturk2024denomamba}, and RWKV \cite{peng2023rwkv,yang2025restorerwkv}—single-task MedIR methods have made substantial progress. However, these single-task models often suffer large performance drops when applied to other MedIR tasks, which limits their practical applicability in broader contexts such as multi-modal imaging scenarios.

\noindent
\textbf{All-in-One Medical Image Restoration.} All-in-One image restoration~\cite{valanarasu2022transweather,li2022airnet,yang2024amir,potlapalli2023promptir,cui2025adair,yang2025tat,chen2025endoir}. aims to address multiple degradation types and modalities using a single unified model. Early attempts in computer vision, such as TransWeather \cite{valanarasu2022transweather}, relied on task-specific encoder–decoder heads to handle distinct weather conditions, inevitably increasing parameter counts as tasks multiplied. To achieve parameter-efficient unified modeling, AirNet \cite{li2022airnet} introduced a contrastive learning approach to generate task-specific latent representations, which serve as prompts to guide a shared restoration network. This prompt-based paradigm has become the dominant strategy, with subsequent methods like PromptIR \cite{potlapalli2023promptir}, AdaIR \cite{cui2025adair}, and others \cite{yang2025tat,chen2025endoir} proposing various mechanisms to learn and inject discriminative prompts for effective task adaptation. In the medical domain, research on All-in-One frameworks is still in its nascent stage~\cite{yang2024amir,yang2025tat,chen2025diffcode,chen2025endoir}. AMIR \cite{yang2024amir} represents a pioneering effort, utilizing a mixture-of-experts strategy to adapt to three specific medical restoration tasks. While these methods have successfully demonstrated the feasibility of unified restoration, they predominantly focus on modeling inter-task heterogeneity—i.e., distinguishing between different tasks to apply specific processing. Consequently, they largely neglect two critical aspects: the inherent homogeneity of anatomical structures shared across medical modalities, and the fine-grained intra-task heterogeneity arising from variations in scanners and protocols. In contrast, our work formulates a hierarchical learning paradigm that jointly models intra- and inter-task homogeneity and heterogeneity, offering a unified perspective for all-in-one  MedIR.

\begin{figure*}[t]
\centering
\includegraphics[width=\textwidth]{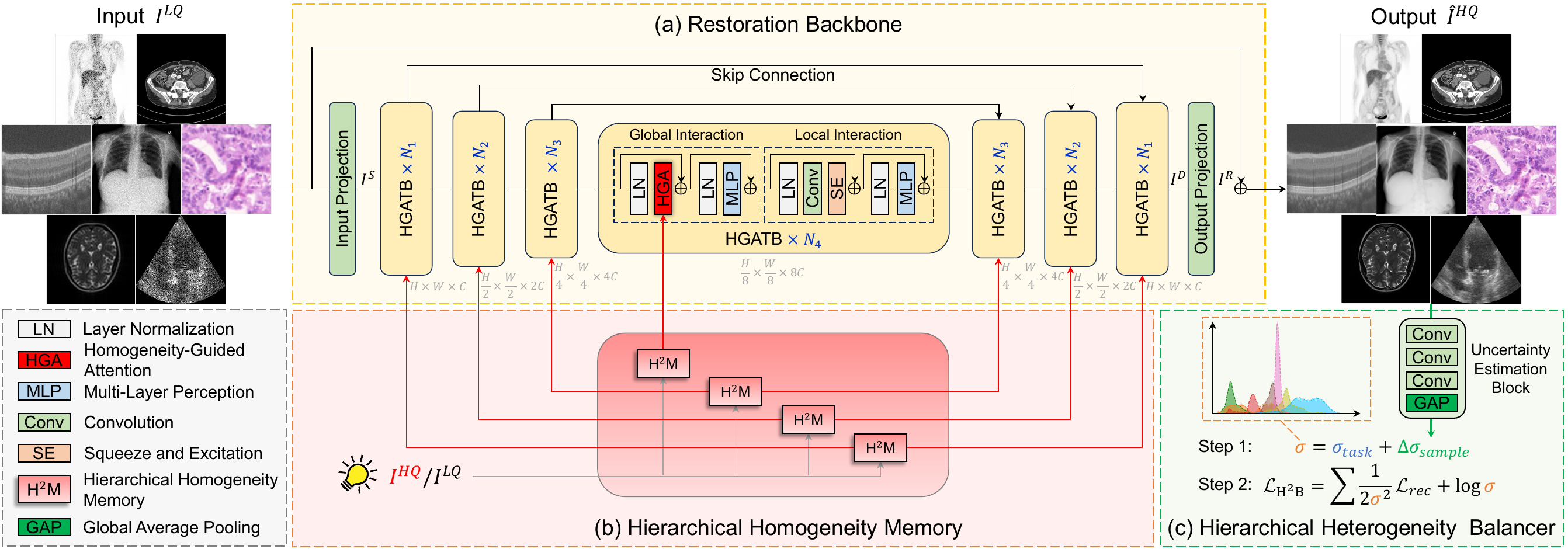}
\caption{The framework of UniH$^3$.
} 
\label{fig_framework}
\end{figure*} 

\section{Method}
Fig.~\ref{fig_framework} illustrates the UniH$^3$ pipeline for all-in-one medical image restoration. UniH$^3$ comprises three main components: a U-shaped restoration backbone (see Fig.~\ref{fig_framework} (a)) responsible for the basic feature extraction and reconstruction, a Hierarchical Homogeneity Memory (H$^2$M, Fig.~\ref{fig_framework} (b)) module that employs both intra- and inter-task homogeneity priors to guide the restoration, and a Hierarchical Heterogeneity Balancer (H$^2$B, see Fig.~\ref{fig_framework} (c)) addresses intra- and inter-task heterogeneity by dynamically balancing task relationships during training. Given a LQ input image $I^{LQ}\in\mathbb{R}^{H\times W\times 1}$, UniH$^3$ first applies a $3\times3$ convolutional input projection to produce shallow features $I^{S}\in\mathbb{R}^{H\times W\times C}$, where $H\times W$ denotes the spatial dimensions and $C$ the number of channels. $I^{S}$ is then processed by a 4-level asymmetric encoder–decoder and transformed into deep features $I^{D}\in\mathbb{R}^{H\times W\times C}$. Each encoder–decoder level contains multiple Homogeneity-Guided Transformer Blocks (HGATBs, see Fig.~\ref{fig_framework} (a)) that extract features under the guidance of H$^2$M-generated priors. Considering that both global and local information are important for medical image restoration \cite{guo2024mambair,yang2025restorerwkv}, each HGATB contains two consecutive transformer-layer variants: the first replaces standard self-attention \cite{dosovitskiy2020vit} with a Homogeneity-Guided Attention (HGA) to model global interactions, and the second replaces self-attention with a convolution plus Squeeze-and-Excitation (SE) \cite{hu2018squeeze_and_excitation} layer to capture local interactions. Finally, $I^{D}$ is projected to a residual image $I^{R}\in\mathbb{R}^{H\times W\times 1}$ by a $3\times3$ convolution, and the restored HQ output is obtained via the residual connection $\hat{I}^{HQ}=I^{LQ}+I^{R}$. We next introduce our core innovations: the H$^2$M module (Sec.~\ref{sec_H$^2$M}), the HGA mechanism (Sec.~\ref{sec_HGA}), and the H$^2$B strategy (Sec.~\ref{sec_H$^2$B}).

\subsection{Hierarchical Homogeneity Memory}
\label{sec_H$^2$M} 
\begin{figure}[t]
  \centering
  \begin{minipage}[b]{0.5\textwidth}
    \centering
    \includegraphics[height=3.5cm]{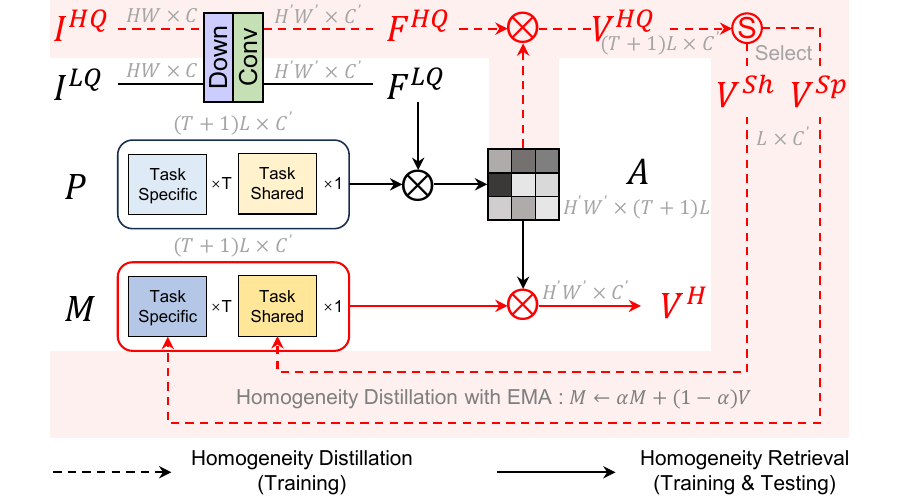}
    \captionof{figure}{Hierarchical Homogeneity Memory.}
    \label{fig_h2m}
  \end{minipage}\hfill
  \begin{minipage}[b]{0.5\textwidth}
    \centering
    \includegraphics[height=3.5cm]{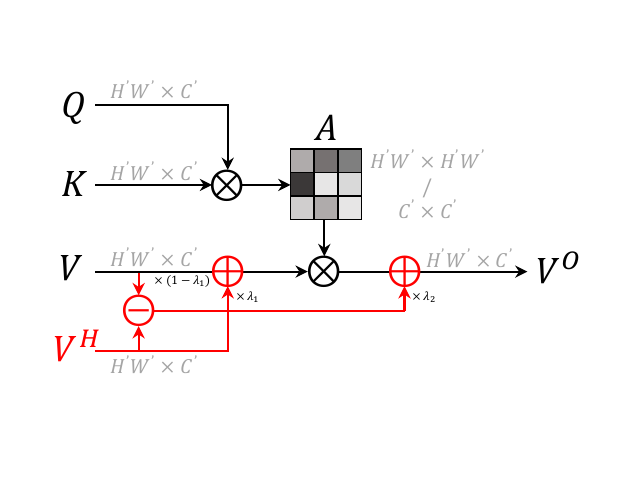}
    \captionof{figure}{Homogeneity-Guided Attention.}
    \label{fig_hga}
  \end{minipage}
\end{figure}

We propose the Hierarchical Homogeneity Memory (H$^2$M) to alleviate the escalating learning difficulty associated with the growing number of restoration tasks and imaging domains. Drawing inspiration from multi-task learning \cite{caruana1997multitasklearning_mtl}, which leverages shared knowledge to reduce learning burdens and accelerate convergence, we observe that HQ medical images exhibit rich homogeneous priors at two hierarchical levels: \textbf{intra-task homogeneity} (i.e., consistent anatomical structures among varying patients within the same modality) and \textbf{inter-task homogeneity} (i.e., shared structured representations of the human anatomy across different imaging modalities). To explicitly model these hierarchical properties, H$^2$M establishes two structurally symmetric components: a memory bank $M \in\mathbb{R}^{(T+1)L\times C'}$ and a learnable prototype matrix $P \in\mathbb{R}^{(T+1)L\times C'}$. Both are organized into $T$ task-specific slots and one task-shared slot (each length of $L$), as illustrated in Fig.~\ref{fig_h2m}. While $M$ is updated via momentum to store distilled HQ anatomical priors, $P$ is a set of learnable parameters that serves as an addressing mechanism,  learning how to optimally store and retrieve information from $M$. The H$^2$M mechanism operates in two phases: Homogeneity Distillation and Homogeneity Retrieval.

\noindent
\textbf{Homogeneity Distillation.} To acquire compact homogeneity priors that facilitate all-in-one restoration, we distill clean anatomical structures from HQ medical images and progressively archive them into $M$ using an Exponential Moving Average (EMA) during training. Concretely, we project paired LQ–HQ images $I^{LQ}$, $I^{HQ}$ to the target resolution via pixel-unshuffle downsampling followed by a $3\times3$ convolution, obtaining paired features $F^{LQ}, F^{HQ} \in\mathbb{R}^{H'W'\times C'}$. A learnable prototype $P\in\mathbb{R}^{L\times C'}$ with the length of $L$ is then used to query and aggregate crucial HQ priors from $F^{HQ}$ through cross-attention:
\begin{equation} 
\operatorname{CrossAttention}(Q, K, V) = \operatorname{Softmax}(\frac{QK^{\mathsf{T}}}{\sqrt{C'}})V,
\end{equation}
\begin{equation}
V^{HQ} = \operatorname{CrossAttention}(P, F^{LQ}, F^{HQ}),
\end{equation}
where $V^{HQ}\in\mathbb{R}^{(T+1)L\times C'}$. This operation allows $P$ to learn which HQ features are most representative of the clean anatomical structures. Based on the current task index, we select features $V^{Sh} \in\mathbb{R}^{L\times C'}$ and $V^{Sp} \in\mathbb{R}^{L\times C'}$ from $V^{HQ}$, which are correspondingly stored into the task-shared slot (to store inter-task homogeneity prior) and task-specific slot (to store intra-task homogeneity prior) of $M$ via an EMA strategy:
\begin{equation}
M_{Sh/Sp} \leftarrow \alpha M_{Sh/Sp} + (1-\alpha) V^{Sh/Sp},
\end{equation}
where $\alpha$ is the momentum coefficient, and $M_{Sh/Sp}$ denotes the corresponding shared or specific slots in $M$. Initialized as zero, $M$ gradually accumulates generalized intra- and inter-task homogeneity priors from continuous training batches. Note that this distillation procedure (indicated by dashed red arrows in Fig.~\ref{fig_h2m}) is performed only during training and discarded at test time.

\noindent
\textbf{Homogeneity Retrieval.} Once the hierarchical memory $M$ is updated, we retrieve clean homogeneity priors $V^{H}$ tailored to the LQ input by using the LQ feature $F^{LQ}$ as a query to retrieve the relevant clean prior from the memory $M$ via cross attention:
\begin{equation}
V^{H} = \operatorname{CrossAttention}(F^{LQ}, P, M),
\end{equation}
where $V^{H}\in\mathbb{R}^{H'W'\times C'}$ is the retrieved homogeneity prior. The red arrows in Fig.~\ref{fig_h2m} illustrate the HQ information flow from $I^{HQ}$ through $M$ into the resulting homogeneity prior $V^{H}$. Because the obtained $V^{H}$ is derived from the distilled HQ memory, it is well-suited to compensate for degraded or missing anatomical information in the LQ features.

To facilitate multi-scale guidance, UniH$^3$ incorporates four H$^2$M modules (see Fig.~\ref{fig_framework}(b)) at different levels of the U-shaped restoration backbone so that the retrieved homogeneity priors provide effective restoration guidance across multiple resolutions.

\subsection{Homogeneity-Guided Attention}
\label{sec_HGA} 
To guide the restoration process using homogeneity priors, we propose a novel Homogeneity-Guided Attention (HGA) mechanism. Existing methods typically incorporate restoration guidance via Spatial Feature Transformations (SFT) \cite{wang2018sft} or cross-attention \cite{cui2025adair}, which treat the LQ features as the basis and the guidance features as supplementary. In contrast, HGA fundamentally shifts the learning paradigm: it anchors the learning starting point on the HQ homogeneity priors rather than the degraded LQ features, thereby substantially reducing the learning difficulty and facilitate model convergence. The design of HGA is detailed below.

HGA is highly flexible and can be implemented on either standard self-attention or transposed self-attention. For clarity of exposition, we formulate it here using standard self-attention. Let the query, key, and value be $Q,K,V\in\mathbb{R}^{H'W'\times C'}$, the conventional self-attention output $V^{O}_0$ is
\begin{equation}
V^{O}_0=AV, \qquad A=\operatorname{Softmax}(\frac{QK^{\mathsf{T}}}{\sqrt{C'}}).
\label{eq_attn}
\end{equation} 

To incorporate guidance from the homogeneity prior $V^{H}$, a straightforward variant is to complement the LQ value $V$ with clean $V^{H}$ by direct addition:
\begin{equation}
\begin{aligned}
V^{O}_1=A(V+V^{H})=AV+AV^{H}.
\end{aligned}
\label{eq_residual_value}
\end{equation} 

In Eq.~\ref{eq_residual_value}, the attention mechanism treats $V$ and $V^{H}$ symmetrically. However, the homogeneity prior $V^{H}$ contains higher-fidelity information than the degraded observation $V$, the attention mechanism should preferentially exploit the more reliable $V^{H}$. To encourage such a preference, we introduce a second variant that biases attention away from the LQ value $V$ and toward the homogeneity prior $V^{H}$ by adding identity-based terms to the attention map $A$:
\begin{equation}
\begin{aligned}
V^{O}_2&=(A-I)V+(A+I)V^{H}\\
&=A(V+V^{H})+V^{H}-V,
\end{aligned}
\label{eq_HGA_init}
\end{equation} 
where $I$ denotes the identity matrix. The $\pm I$ terms increase the self-contribution of $V^{H}$ while reducing that of $V$. $V+V^H$ denotes the mixed value, and $V^H-V$ acts as a preference bias that reinforces more reliance on the homogeneity prior $V^H$. To stabilize training and increase model expressivity, the final HGA mechanism is obtained by applying channel-wise learnable weighting parameters $\lambda_1, \lambda_2 \in\mathbb{R}^{ C'}$:
\begin{equation}
V^{O}=A[(1-\lambda_1)V+\lambda_1V^{H}]+\lambda_2(V^{H}-V).
\label{eq_HGA}
\end{equation} 
When $\lambda_1=\lambda_2=0$, the HGA reduces to the conventional self-attention. The self-attention–based HGA in Eq.~\ref{eq_HGA} has an analogous form to transposed self-attention (see \textit{supplement}). Our proposed UniH$^3$ adopts the HGA based on transposed self-attention following Restormer \cite{zamir2022restormer}. Fig.~\ref{fig_hga} illustrates the HGA formulation, which augments attention with simple addition and subtraction operations on the value.

\subsection{Hierarchical Heterogeneity Balancer}
\label{sec_H$^2$B}
We propose a Hierarchical Heterogeneity Balancer (H$^2$B) to mitigate inter- and intra-task heterogeneity across diverse MedIR tasks during the optimization process. Heterogeneity among tasks induces gradient conflicts that create an imbalance in optimization: some tasks dominate training while others remain under-trained. Previous work in multi-task learning \cite{kendall2018uncertainty_loss} and all-in-one natural image restoration \cite{wu2025debiased_uncertainty_loss} has shown that uncertainty-based loss balancing is a good way of addressing inter-task heterogeneity by dynamically scaling different task losses for a reasonable optimization route:
\begin{equation}
\mathcal{L}_{\text {UB}}=\frac{1}{T}\sum_{t=1}^{T}\left(\frac{1}{2 \sigma_{t}^2} \mathcal{L}_{rec}^{(t)}+\log \sigma_{t}\right),
\end{equation} 
where $T$ denotes the number of tasks, $\mathcal{L}_{rec}^{(t)}$ denotes the reconstruction loss, and $\sigma_t$ is a learnable scalar that estimates task-level uncertainty. The factor $\frac{1}{2\sigma _{t}^{2}}$ adaptively rescales each task’s contribution while the $\log\sigma_t$ term regularizes the scaling. When $\mathcal{L}_{rec}^{(t)}$ increases and tends to dominate the total loss, $\sigma_t$ increases to attenuate its contribution, and vice versa. However, this uncertainty balancing is too coarse: a single scalar $\sigma_t$ per-task cannot capture intra-task heterogeneity (e.g., scanner/center/anatomy variations), so hard samples still remain insufficiently handled. We therefore introduce a hierarchical uncertainty model. For task $t$ and sample $s$ we define the total uncertainty $\sigma_{t,s}$ as the sum of a global task term $\sigma_t$ and a sample-specific correction $\Delta\sigma_{s}$:
\begin{equation}
\sigma_{t, s}=\sigma_t+\Delta \sigma_s,
\end{equation} 
where $t$ indexes tasks and $s$ indexes samples. $\sigma_t$ is still a learnable scalar for each task while $\Delta\sigma_s$ is predicted by a lightweight Uncertainty Estimation Block (UEB, see Fig.~\ref{fig_framework}(c)) conditioned on sample-specific signals:
\begin{equation}
\Delta \sigma_s=\operatorname{UEB}\left(\operatorname{Concat}[I_s^{LQ},\operatorname{sg}(\hat{I}_s^{HQ}),I_s^{HQ}]\right),
\label{eq_UEB}
\end{equation}
where $I_s^{LQ}$ is the LQ input for sample $s$, $\hat I_s^{HQ}$ is the model prediction, $I_s^{HQ}$ is the HQ ground truth, and $\operatorname{sg}(\cdot)$ denotes stop-gradient to decouple loss balancing from restoration model optimization. The H$^2$B loss then aggregates per-task and per-sample contributions as:
\begin{equation}
\mathcal{L}_{\text {H$^2$B}}=\frac{1}{T S}\sum_{t=1}^{T}\sum_{s=1}^{S} \left(\frac{1}{2 \sigma_{t,s}^2} \mathcal{L}_{rec}^{(t,s)}+\log \sigma_{t,s}\right),
\end{equation}
where $S$ is the batch size. H$^2$B retains the theoretical foundation of standard uncertainty-based balancing \cite{kendall2018uncertainty_loss} while refining it to capture uncertainty at two hierarchical levels: a task-level term $\sigma_t$ to effectively mitigate inter-task heterogeneity, and a sample-level correction $\Delta\sigma_s$ to mitigate intra-task heterogeneity.

\section{Experiments} 
We conduct experiments under two settings, \textit{All-in-One} and \textit{Single-Task}, for both 2D and 3D MedIR tasks. In the \textit{All-in-One} setting, a single universal model is trained to address multiple MedIR tasks within either the 2D or 3D domain. In the \textit{Single-Task} setting, separate models are trained for each MedIR task. We first describe the experimental setup, including datasets, implementation details, and evaluation. We then present comparative results in Sec.~\ref{sec_result_all_in_one} and Sec.~\ref{sec_result_single}, and ablation studies in Sec.~\ref{sec_ablation}.

\noindent
\textbf{Datasets.} Most existing MedIR datasets are limited in size and narrowly tailored to specific tasks and modalities. To promote the development of general-purpose MedIR methods, we organize publicly available datasets together with private collections into two datasets, as summarized in Tab.~\ref{tab_dataset}: \textbf{(i) MedIR-2D-500K} comprises 509,200 2D LQ-HQ image pairs across seven distinct 2D MedIR tasks: PET image denoising, CT image denoising, MRI image super-resolution, X-ray image denoising, OCT image denoising, ultrasound image denoising, and pathological image super-resolution. \textbf{(ii) MedIR-3D-3K} includes 3,522 3D LQ-HQ volume pairs covering three 3D MedIR tasks: PET image denoising, CT image denoising, and MRI image super-resolution. We expect these two datasets to serve as a useful benchmark for advancing general-purpose MedIR research. More detailed descriptions are shown in the \textit{supplement}.

\begin{table}[!t]
\caption{Overview of the MedIR-2D-500K and MedIR-3D-3K datasets.}
\centering
\resizebox{\textwidth}{!}{
\begin{tabular}{ccccccc}
\toprule
\textbf{Dataset}                & \textbf{Dimension}   & \textbf{Modality}                      & \textbf{Training}                       & \textbf{Testing}                       & \textbf{Total}                          & \textbf{Data Source}                        \\ \midrule
                                &                      & PET                                    & 77,000                                   & 8,600                                   & 85,600                                   & \cite{xue2022cross}, Private \\
                                &                      & CT                                     & 64,000                                   & 8,100                                   & 72,100                                   & \cite{mccollough2017aapm,moen2021ldct} , Private  \\
                                &                      & MRI                                    & 75,000                                   & 8,400                                   & 83,400                                   & \cite{van2013HCP,ixi_dataset}  \\
                                &                      & X-ray                                  & 108,000                                  & 11,600                                  & 119,600                                  & \cite{wang2017chestxray,chowdhury2020covid,rahman2021exploring,wang2023medfmc}  \\
                                &                      & OCT                                    & 32,000                                   & 3,600                                   & 35,600                                   & \cite{li2024octa,geng2022pku,fang2012sbsdi}  \\
                                &                      & Ultrasound                             & 56,000                                   & 5,800                                   & 61,800                                   & \cite{yiguo2023usenhance,ultrasound-nerve-segmentation,van2018hc,leclerc2019camus,yang2023cardiac} \\
                                &                      & Pathology                              & 46,000                                   & 5,100                                   & 51,100                                   & \cite{drifka2016tma,kumar2019monuseg,da2022digestpath,sirinukunwattana2017glas,aksac2019brecahad,tekin2023tubule} \\
\multirow{-8}{*}{\textbf{MedIR-2D-500K}} & \multirow{-8}{*}{2D} & \cellcolor[HTML]{EFEFEF}\textbf{Total} & \cellcolor[HTML]{EFEFEF}\textbf{458,000} & \cellcolor[HTML]{EFEFEF}\textbf{51,200} & \cellcolor[HTML]{EFEFEF}\textbf{509,200} & \cellcolor[HTML]{EFEFEF}-                   \\ \hline
                                &                      & PET                                    & 1,388                                    & 156                                    & 1,544                                    & \cite{xue2022cross}, Private \\
                                &                      & CT                                     & 258                                     & 30                                     & 288                                     & \cite{mccollough2017aapm,moen2021ldct}, Private \\
                                &                      & MRI                                    & 1,520                                    & 170                                    & 1,690                                    & \cite{van2013HCP,ixi_dataset} \\
\multirow{-4}{*}{\textbf{MedIR-3D-3K}}   & \multirow{-4}{*}{3D} & \cellcolor[HTML]{EFEFEF}\textbf{Total} & \cellcolor[HTML]{EFEFEF}\textbf{3,166}   & \cellcolor[HTML]{EFEFEF}\textbf{356}   & \cellcolor[HTML]{EFEFEF}\textbf{3,522}   & \cellcolor[HTML]{EFEFEF}-                   \\ \bottomrule
\end{tabular}
} 
\label{tab_dataset}
\end{table} 

\noindent
\textbf{Implementation.} For the UniH$^3$ architecture, the number of HGATBs are $N_1=2$, $N_2=N_3=3$, and $N_4=4$. The input channel dimension is $C=48$. For the H$^2$M module, we set the the number of tasks $T=7$, memory length $L=128$ and EMA coefficient $\alpha=0.99$. During training we use patches of size $128\times128$ with a batch size of 14. The reconstruction loss $\mathcal{L}_{rec}$ is defined as the L1 loss. The model is optimized using Muon optimizer~\cite{liu2025muon} for $6\times10^{5}$ iterations, with an initial learning rate $3\times10^{-4}$ and annealed to $1\times10^{-7}$ using a cosine schedule. To support 3D MedIR, we introduce a 3D variant, UniH$^3$-3D, obtained by replacing each module in UniH$^3$ with its 3D counterpart. For UniH$^3$-3D, the HGATB numbers are $N_1=N_2=1$ and $N_3=N_4=5$, the input channel number is $C=16$, and the init learning rate is set as $5\times10^{-5}$. The patch size is $64\times64\times64$ and the batch size is 6. All other settings are identical to the original UniH$^3$.

\noindent
\textbf{Evaluation.} To quantitatively assess image quality, we employ the widely used PSNR and SSIM metrics.  In the reported tables, the highest and second-highest scores are indicated in \textcolor{red}{red} and \textcolor{blue}{blue}, respectively.

\begin{table*}[!t]
\caption{All-in-one MedIR comparison results on the MedIR-2D-500K dataset.}
\centering
\resizebox{\textwidth}{!}{
\begin{tabular}{ccccccccccccccccccccccccccccc}
\toprule
                                  &  &                                           &                      &                                      &  & \multicolumn{2}{c}{\textbf{PET}}                                               &                                  & \multicolumn{2}{c}{\textbf{CT}}                                                &                                  & \multicolumn{2}{c}{\textbf{MRI}}                                               &                                  & \multicolumn{2}{c}{\textbf{X-ray}}                                             &                                  & \multicolumn{2}{c}{\textbf{OCT}}                                               &                                  & \multicolumn{2}{c}{\textbf{Ultrasound}}                                        &                                  & \multicolumn{2}{c}{\textbf{Pathology}}                                         &                                  & \multicolumn{2}{c}{\textbf{Average}}                                           \\ \cline{7-8} \cline{10-11} \cline{13-14} \cline{16-17} \cline{19-20} \cline{22-23} \cline{25-26} \cline{28-29} 
\multirow{-2}{*}{\textbf{Method}} &  & \multirow{-2}{*}{\textbf{\#Params   (M)}} &                      & \multirow{-2}{*}{\textbf{FLOPs (G)}} &  & \textbf{PSNR↑}                        & \textbf{SSIM↑}                         &                                  & \textbf{PSNR↑}                        & \textbf{SSIM↑}                         &                                  & \textbf{PSNR↑}                        & \textbf{SSIM↑}                         &                                  & \textbf{PSNR↑}                        & \textbf{SSIM↑}                         &                                  & \textbf{PSNR↑}                        & \textbf{SSIM↑}                         &                                  & \textbf{PSNR↑}                        & \textbf{SSIM↑}                         &                                  & \textbf{PSNR↑}                        & \textbf{SSIM↑}                         &                                  & \textbf{PSNR↑}                        & \textbf{SSIM↑}                         \\ \midrule
SwinIR \cite{liang2021swinir}                           &  & 11.50                                     &                      & 187.93                               &  & 44.24                                 & 0.9866                                 &                                  & 43.17                                 & 0.9338                                 &                                  & 38.44                                 & 0.9453                                 &                                  & 35.87                                 & 0.9275                                 &                                  & 35.70                                 & 0.8892                                 &                                  & 27.52                                 & 0.8089                                 &                                  & 28.35                                 & 0.7941                                 &                                  & 36.18                                 & 0.8979                                 \\
Uformer \cite{wang2022uformer}                      &  & 50.47                                     &                      & 21.42                                &  & 44.51                                 & 0.9874                                 &                                  & 43.45                                 & 0.9356                                 &                                  & 39.01                                 & 0.9507                                 &                                  & 36.59                                 & {\color[HTML]{3531FF}0.9353}                                 &                                  & 35.84                                 & 0.8908                                 &                                  & 27.61                                 & 0.8135                                 &                                  & 28.55                                 & 0.8011                                 &                                  & 36.51                                 & 0.9020                                 \\
Restormer \cite{zamir2022restormer}                      &  & 26.12                                     &                      & 35.21                                &  & 44.47                                 & 0.9873                                 &                                  & 43.44                                 & 0.9353                                 &                                  & 39.05                                 & 0.9515                                 &                                  & 36.46                                 & 0.9335                                 &                                  & 35.84                                 & 0.8909                                 &                                  & 27.66                                 & 0.8138                                 &                                  & 28.53                                 & 0.8006                                 &                                  & 36.49                                 & 0.9018                                 \\
NAFNet \cite{chen2022nafnet}                            &  & 67.89                                     &                      & 15.74                                &  & 44.40                                 & 0.9871                                 &                                  & 43.32                                 & 0.9346                                 &                                  & 38.90                                 & 0.9502                                 &                                  & 36.33                                 & 0.9326                                 &                                  & 35.82                                 & 0.8905                                 &                                  & 27.59                                 & 0.8131                                 &                                  & 28.49                                 & 0.7995                                 &                                  & 36.41                                 & 0.9011                                 \\
Restore-RWKV   \cite{yang2025restorerwkv}                   &  & 27.91                                     &                      & 37.46                                &  & 44.45                                 & 0.9872                                 &                                  & 43.46                                 & 0.9352                                 &                                  & 39.05                                 & 0.9514                                 &                                  & 36.46                                 & 0.9333                                 &                                  & 35.84                                 & 0.8909                                 &                                  & 27.68                                 & 0.8149                                 &                                  & 28.55                                 & 0.8011                                 &                                  & 36.50                                 & 0.9020                                 \\
MambaIR \cite{guo2024mambair}                          &  & 31.50                                     &                      & 34.35                                &  & 44.50                                 & 0.9873                                 &                                  & 43.47                                 & 0.9355                                 &                                  & 39.07                                 & 0.9517                                 &                                  & 36.57                                 & 0.9341                                 &                                  & 35.83                                 & 0.8904                                 &                                  & 27.69                                 & 0.8146                                 &                                  & 28.53                                 & 0.8004                                 &                                  & 36.52                                 & 0.9020                                 \\ \hline
TransWeather  \cite{valanarasu2022transweather}                    &  & 38.05                                     & \multicolumn{1}{l}{} & 1.55                                 &  & 43.73                                 & 0.9846                                 &                                  & 41.12                                 & 0.9217                                 &                                  & 37.62                                 & 0.9333                                 &                                  & 35.28                                 & 0.9221                                 &                                  & 35.12                                 & 0.8699                                 &                                  & 27.23                                 & 0.8011                                 &                                  & 28.17                                 & 0.7874                                 &                                  & 35.47                                 & 0.8886                                 \\
AirNet \cite{li2022airnet}                           &  & 7.61                                      &                      & 230.48                               &  & 44.32                                 & 0.9868                                 &                                  & 43.34                                 & 0.9348                                 &                                  & 38.81                                 & 0.9489                                 &                                  & 36.38                                 & 0.9322                                 &                                  & 35.77                                 & 0.8900                                 &                                  & 27.62                                 & 0.8127                                 &                                  & 28.46                                 & 0.7981                                 &                                  & 36.39                                 & 0.9005                                 \\
DRMC \cite{yang2023drmc}                             &  & 0.62                                      &                      & 9.92                                 &  & 43.62                                 & 0.9841                                 &                                  & 42.48                                 & 0.9281                                 &                                  & 37.24                                 & 0.9301                                 &                                  & 34.06                                 & 0.9085                                 &                                  & 35.46                                 & 0.8862                                 &                                  & 27.09                                 & 0.7993                                 &                                  & 28.02                                 & 0.7830                                 &                                  & 35.42                                 & 0.8885                                 \\
AMIR  \cite{yang2024amir}                            &  & 23.54                                     &                      & 31.76                                &  & 44.49                                 & 0.9873                                 &                                  & 43.47                                 & 0.9356                                 &                                  & 39.09                                 & 0.9519                                 &                                  & 36.47                                 & 0.9333                                 &                                  &{\color[HTML]{3531FF} 35.89}                                  & 0.8914                                 &                                  & 27.69                                 & 0.8150                                 &                                  & {\color[HTML]{3531FF} 28.57}                                 & {\color[HTML]{3531FF} 0.8019}                                 &                                  & 36.52                                 & 0.9023                                 \\
PromptIR \cite{potlapalli2023promptir}                         &  & 35.59                                     &                      & 39.49                                &  & 44.52                                 & 0.9874                                 &                                  & 43.48                                 & 0.9355                                 &                                  & 39.13                                 & 0.9524                                 &                                  & 36.57                                 & 0.9341                                 &                                  & 35.84                                 & 0.8909                                 &                                  & {\color[HTML]{3531FF} 27.69}                                 & {\color[HTML]{3531FF} 0.8152}                                 &                                  & 28.54                                 & 0.8006                                 &                                  & 36.54                                 & 0.9023                                 \\
AdaIR \cite{cui2025adair}                            &  & 28.76                                     &                      & 36.74                                &  & {\color[HTML]{3531FF} 44.55}          & {\color[HTML]{3531FF} 0.9875}          &                                  & {\color[HTML]{3531FF} 43.49}          & {\color[HTML]{3531FF} 0.9356}          &                                  & {\color[HTML]{3531FF} 39.17}          & {\color[HTML]{3531FF} 0.9527}          &                                  & {\color[HTML]{3531FF} 36.60}          & 0.9344          &                                  & 35.86          & {\color[HTML]{3531FF} 0.8915}          &                                  & 27.69          & 0.8150          &                                  & 28.56          & 0.8015          &                                  & {\color[HTML]{3531FF} 36.56}          & {\color[HTML]{3531FF} 0.9026}          \\\rowcolor[HTML]{EFEFEF}
\textbf{UniH$^3$ (Ours)}            &  & 28.96                                     &                      & 26.33                                &  & {\color[HTML]{FF0000} \textbf{44.89}} & {\color[HTML]{FF0000} \textbf{0.9883}} & {\color[HTML]{FF0000} \textbf{}} & {\color[HTML]{FF0000} \textbf{43.65}} & {\color[HTML]{FF0000} \textbf{0.9368}} & {\color[HTML]{FF0000} \textbf{}} & {\color[HTML]{FF0000} \textbf{39.55}} & {\color[HTML]{FF0000} \textbf{0.9564}} & {\color[HTML]{FF0000} \textbf{}} & {\color[HTML]{FF0000} \textbf{36.88}} & {\color[HTML]{FF0000} \textbf{0.9368}} & {\color[HTML]{FF0000} \textbf{}} & {\color[HTML]{FF0000} \textbf{35.96}} & {\color[HTML]{FF0000} \textbf{0.8921}} & {\color[HTML]{FF0000} \textbf{}} & {\color[HTML]{FF0000} \textbf{27.80}} & {\color[HTML]{FF0000} \textbf{0.8179}} & {\color[HTML]{FF0000} \textbf{}} & {\color[HTML]{FF0000} \textbf{28.63}} & {\color[HTML]{FF0000} \textbf{0.8035}} & {\color[HTML]{FF0000} \textbf{}} & {\color[HTML]{FF0000} \textbf{36.77}} & {\color[HTML]{FF0000} \textbf{0.9045}} \\ \bottomrule
\end{tabular}
}
\label{tab_all_in_one_2D}
\end{table*} 

\begin{table*}[!t]
\caption{Single-task MedIR comparison results on the MedIR-2D-500K dataset.}
\centering
\resizebox{\textwidth}{!}{
\begin{tabular}{ccccccccccccccccccccccccccccc}
\toprule
                                  &  &                                         &  &                                      &  & \multicolumn{2}{c}{\textbf{PET}}                                               &  & \multicolumn{2}{c}{\textbf{CT}}                                                &  & \multicolumn{2}{c}{\textbf{MRI}}                                               &  & \multicolumn{2}{c}{\textbf{X-ray}}                                             &  & \multicolumn{2}{c}{\textbf{OCT}}                                               &  & \multicolumn{2}{c}{\textbf{Ultrasound}}                                        &  & \multicolumn{2}{c}{\textbf{Pathology}}                                         &  & \multicolumn{2}{c}{\textbf{Average}}                                           \\ \cline{7-8} \cline{10-11} \cline{13-14} \cline{16-17} \cline{19-20} \cline{22-23} \cline{25-26} \cline{28-29} 
\multirow{-2}{*}{\textbf{Method}} &  & \multirow{-2}{*}{\textbf{\#Params (M)}} &  & \multirow{-2}{*}{\textbf{FLOPs (G)}} &  & \textbf{PSNR↑}                        & \textbf{SSIM↑}                         &  & \textbf{PSNR↑}                        & \textbf{SSIM↑}                         &  & \textbf{PSNR↑}                        & \textbf{SSIM↑}                         &  & \textbf{PSNR↑}                        & \textbf{SSIM↑}                         &  & \textbf{PSNR↑}                        & \textbf{SSIM↑}                         &  & \textbf{PSNR↑}                        & \textbf{SSIM↑}                         &  & \textbf{PSNR↑}                        & \textbf{SSIM↑}                         &  & \textbf{PSNR↑}                        & \textbf{SSIM↑}                         \\ \midrule
SwinIR \cite{liang2021swinir}                           &  & 11.50                                   &  & 187.93                               &  & 44.08                                 & 0.9860                                 &  & 43.53                                 & 0.9358                                 &  & 39.19                                 & 0.9529                                 &  & 36.30                                 & 0.9308                                 &  & 35.80                                 & 0.8904                                 &  & 27.68                                 & 0.8143                                 &  & 28.50                                 & 0.7993                                 &  & 36.44                                 & 0.9014                                 \\
Uformer \cite{wang2022uformer}                          &  & 50.47                                   &  & 21.42                                &  & 44.61                                 & 0.9876                                 &  & 43.54                                 & 0.9362                                 &  & 39.21                                 & 0.9526                                 &  & 36.61                                 & 0.9363                                 &  & 35.91                                 & 0.8915                                 &  & 27.64                                 & 0.8151                                 &  & 28.58                                 & 0.8022                                 &  & 36.59                                 & 0.9031                                 \\
Restormer \cite{zamir2022restormer}                        &  & 26.12                                   &  & 35.21                                &  & 44.90                                 & 0.9883                                 &  & {\color[HTML]{0000FF} 43.64}                                 & {\color[HTML]{0000FF} 0.9369}                                 &  & 39.44                                 & 0.9554                                 &  & 36.71                                 & 0.9359                                 &  & 35.97                                 & 0.8921                                 &  & 27.76                                 & 0.8164                                 &  & 28.63                                 & 0.8038                                 &  & 36.72                                 & 0.9041                                 \\
NAFNet \cite{chen2022nafnet}                            &  & 67.89                                   &  & 15.74                                &  & 44.74                                 & 0.9880                                 &  & 43.55                                 & 0.9362                                 &  & 39.29                                 & 0.9540                                 &  & 36.48                                 & 0.9343                                 &  & 35.93                                 & 0.8919                                 &  & 27.67                                 & 0.8139                                 &  & 28.58                                 & 0.8024                                 &  & 36.61                                 & 0.9030                                 \\
Restore-RWKV \cite{yang2025restorerwkv}                     &  & 27.91                                   &  & 37.46                                &  & {\color[HTML]{0000FF} 44.93}                                 & {\color[HTML]{0000FF} 0.9884}                                 &  & 43.64                                 & 0.9369                                 &  & 39.57                                 & 0.9565                                 &  & 36.77                                 & 0.9359                                 &  & 35.97                                 & 0.8923                                 &  & 27.76                                 & 0.8168                                 &  & 28.62                                 & 0.8036                                 &  & 36.75                                 & 0.9043                                 \\
MambaIR \cite{guo2024mambair}                          &  & 31.50                                   &  & 34.35                                &  & 44.93          & 0.9883          &  &  43.55          & 0.9363          &  & {\color[HTML]{0000FF} 39.59}          & {\color[HTML]{0000FF} 0.9568}          &  & {\color[HTML]{0000FF} 36.77}          & {\color[HTML]{0000FF} 0.9363}          &  & {\color[HTML]{0000FF} 35.99}          & {\color[HTML]{0000FF} 0.8923}          &  & {\color[HTML]{0000FF} 27.77}          & {\color[HTML]{0000FF} 0.8178}          &  & {\color[HTML]{0000FF} 28.64}          & {\color[HTML]{0000FF} 0.8041}          &  & {\color[HTML]{0000FF} 36.75}          & {\color[HTML]{0000FF} 0.9046}          \\\rowcolor[HTML]{EFEFEF}
\textbf{UniH$^3$ (Ours)}                   &  & 28.96                                   &  & 26.33                                &  & {\color[HTML]{FF0000} \textbf{45.11}} & {\color[HTML]{FF0000} \textbf{0.9889}} &  & {\color[HTML]{FF0000} \textbf{43.77}} & {\color[HTML]{FF0000} \textbf{0.9376}} &  & {\color[HTML]{FF0000} \textbf{39.78}} & {\color[HTML]{FF0000} \textbf{0.9584}} &  & {\color[HTML]{FF0000} \textbf{37.05}} & {\color[HTML]{FF0000} \textbf{0.9384}} &  & {\color[HTML]{FF0000} \textbf{36.04}} & {\color[HTML]{FF0000} \textbf{0.8929}} &  & {\color[HTML]{FF0000} \textbf{27.86}} & {\color[HTML]{FF0000} \textbf{0.8192}} &  & {\color[HTML]{FF0000} \textbf{28.69}} & {\color[HTML]{FF0000} \textbf{0.8050}} &  & {\color[HTML]{FF0000} \textbf{36.90}} & {\color[HTML]{FF0000} \textbf{0.9058}} \\ \bottomrule
\end{tabular}
} 
\label{tab_single_2D}
\end{table*}

\begin{figure*}[!t]
\centering
\includegraphics[width=\textwidth]{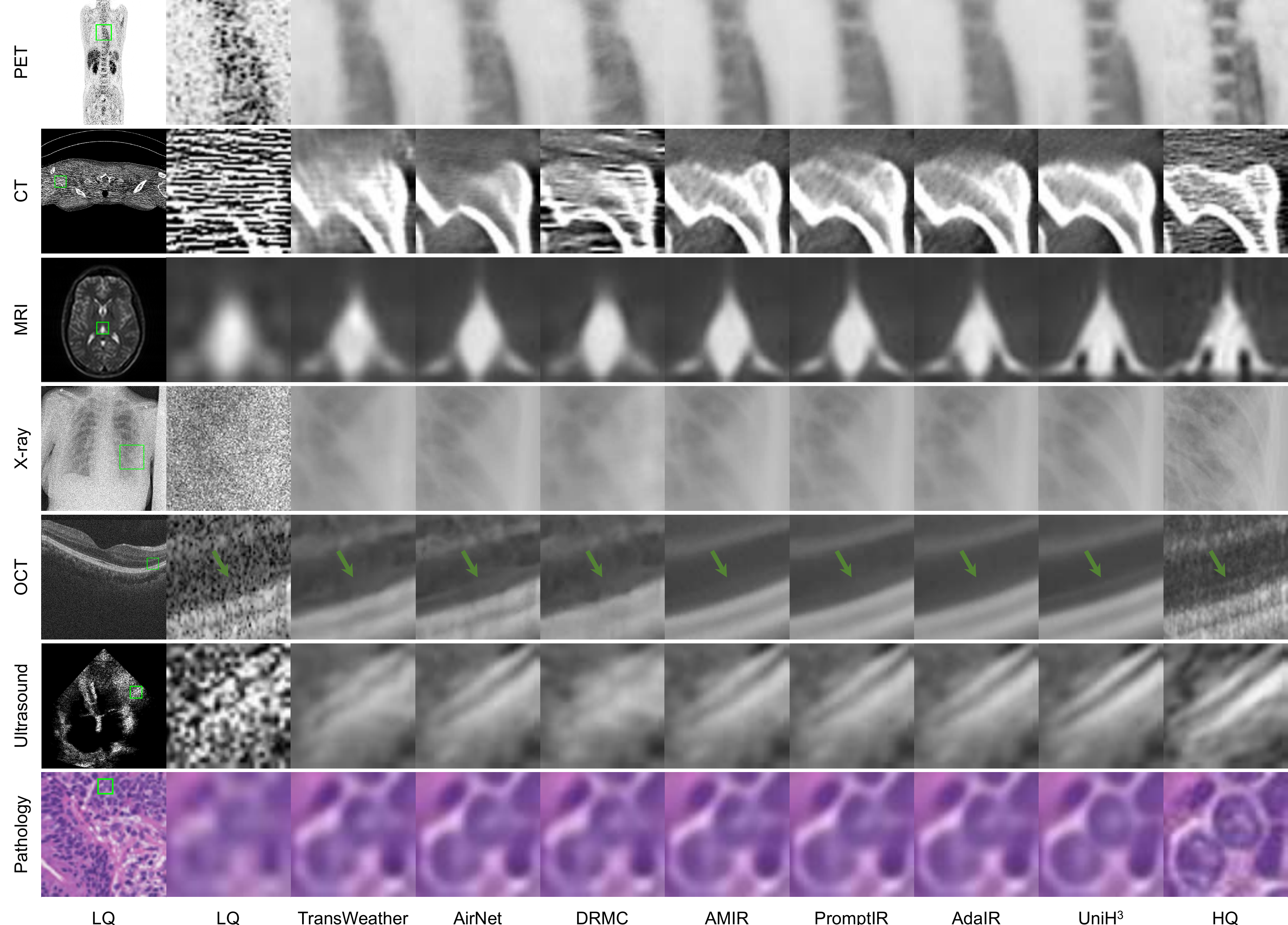}
\caption{Visual comparison of methods for all-in-one medical image restoration on the MedIR-2D-500K dataset.
} 
\label{fig_all_in_one_2D_vis}
\end{figure*}

\begin{table}[!t]
\centering
\begin{minipage}[t]{0.49\textwidth}
\centering
\captionof{table}{3D all-in-one MedIR results on the MedIR-3D-3K dataset.}
\label{tab_all_in_one_3D}
\resizebox{\linewidth}{!}{
\begin{tabular}{ccccccccccccc}
\toprule
                                  &  & \multicolumn{2}{c}{\textbf{PET}}                                               & \textbf{}                        & \multicolumn{2}{c}{\textbf{CT}}                                                & \textbf{}                        & \multicolumn{2}{c}{\textbf{MRI}}                                               & \textbf{}                        & \multicolumn{2}{c}{\textbf{Average}}                                           \\ \cline{3-4} \cline{6-7} \cline{9-10} \cline{12-13} 
\multirow{-2}{*}{\textbf{Method}} &  & \textbf{PSNR↑}                        & \textbf{SSIM↑}                         & \textbf{}                        & \textbf{PSNR↑}                        & \textbf{SSIM↑}                         & \textbf{}                        & \textbf{PSNR↑}                        & \textbf{SSIM↑}                         & \textbf{}                        & \textbf{PSNR↑}                        & \textbf{SSIM↑}                         \\ \midrule
3D-cGAN \cite{wang20183dcgan}                          &  & 48.22                                 & 0.9937                                 &                                  & 40.82                                 & 0.9320                                 &                                  & 38.57                                 & 0.9489                                 &                                  & 42.54                                 & 0.9582                                 \\
MRDG \cite{wang2020mrdg}                              &  & 48.68                                 & 0.9943                                 &                                  & 42.86                                 & 0.9388                                 &                                  & 38.93                                 & 0.9519                                 &                                  & 43.49                                 & 0.9617                                 \\
DRMC \cite{yang2023drmc}                             &  & 48.58                                 & 0.9933                                 &                                  & 43.33                                 & 0.9391                                 &                                  & 38.81                                 & 0.9514                                 &                                  & 43.57                                 & 0.9613                                 \\
Spach Transformer \cite{jang2023spachtransformer}                 &  & 48.91                                 & 0.9949                                 &                                  & 42.21                                 & 0.9380                                 &                                  & 39.04                                 & 0.9542                                 &                                  & 43.39                                 & 0.9624                                 \\
Restore-RWKV-3D \cite{yang2025restorerwkv}                  &  & {\color[HTML]{0000FF} 49.08}           & {\color[HTML]{0000FF} 0.9951}          &                                  & {\color[HTML]{0000FF}43.35}           & {\color[HTML]{0000FF} 0.9402}          &                                  & {\color[HTML]{0000FF} 39.43}           & {\color[HTML]{0000FF} 0.9578}          &                                  & {\color[HTML]{0000FF} 43.95}           & {\color[HTML]{0000FF} 0.9644}          \\\rowcolor[HTML]{EFEFEF}
\textbf{UniH$^3$-3D (Ours)}          &  & {\color[HTML]{FF0000} \textbf{49.43}} & {\color[HTML]{FF0000} \textbf{0.9955}} & {\color[HTML]{FF0000} \textbf{}} & {\color[HTML]{FF0000} \textbf{44.05}} & {\color[HTML]{FF0000} \textbf{0.9429}} & {\color[HTML]{FF0000} \textbf{}} & {\color[HTML]{FF0000} \textbf{40.03}} & {\color[HTML]{FF0000} \textbf{0.9637}} & {\color[HTML]{FF0000} \textbf{}} & {\color[HTML]{FF0000} \textbf{44.50}} & {\color[HTML]{FF0000} \textbf{0.9674}} \\ \bottomrule
\end{tabular}
}
\end{minipage}
\hfill
\begin{minipage}[t]{0.49\textwidth}
\centering
\captionof{table}{3D single-task MedIR results on the MedIR-3D-3K dataset.}
\label{tab_single_3D}
\resizebox{\linewidth}{!}{
\begin{tabular}{ccccccccccccc}
\toprule
                                  &  & \multicolumn{2}{c}{\textbf{PET}}                                               & \textbf{}                        & \multicolumn{2}{c}{\textbf{CT}}                                                & \textbf{}                        & \multicolumn{2}{c}{\textbf{MRI}}                                               & \textbf{}                        & \multicolumn{2}{c}{\textbf{Average}}                                           \\ \cline{3-4} \cline{6-7} \cline{9-10} \cline{12-13} 
\multirow{-2}{*}{\textbf{Method}} &  & \textbf{PSNR↑}                        & \textbf{SSIM↑}                         & \textbf{}                        & \textbf{PSNR↑}                        & \textbf{SSIM↑}                         & \textbf{}                        & \textbf{PSNR↑}                        & \textbf{SSIM↑}                         & \textbf{}                        & \textbf{PSNR↑}                        & \textbf{SSIM↑}                         \\ \midrule
3D-cGAN \cite{wang20183dcgan}                          &  & 48.47                                 & 0.9944                                 &                                  & 40.86                                 & 0.9309                                 &                                  & 38.76                                 & 0.9510                                 &                                  & 42.70                                 & 0.9588                                 \\
MRDG \cite{wang2020mrdg}                             &  & 48.40                                 & 0.9942                                 &                                  & 43.14                                 & 0.9395                                 &                                  & 39.34                                 & {\color[HTML]{0000FF}0.9570}                                &                                  & 43.63                                 & {\color[HTML]{0000FF}0.9636}                                 \\
DRMC \cite{yang2023drmc}                             &  & {\color[HTML]{0000FF} 48.76}                                & {\color[HTML]{0000FF} 0.9947}                                 &                                  & {\color[HTML]{0000FF}43.60}                                & 0.9404                                 &                                  & 38.98                                 & 0.9540                                 &                                  & 43.78                                 & 0.9630                                 \\
Spach Transformer \cite{jang2023spachtransformer}                &  & 48.71                                 & 0.9946                                 &                                  & 43.57                                 & {\color[HTML]{0000FF}0.9409}                                 &                                  & {\color[HTML]{0000FF}39.16}                                 & 0.9546                                 &                                  & {\color[HTML]{0000FF}43.81}                                 & 0.9634                                 \\
Restore-RWKV-3D \cite{yang2025restorerwkv}                  &  & 48.67           &  0.9945          &           &  42.48           &  0.9375          &           &  39.15           &  0.9551          &           & 43.43           & 0.9624          \\\rowcolor[HTML]{EFEFEF}
\textbf{UniH$^3$-3D (Ours)}          &  & {\color[HTML]{FF0000} \textbf{49.77}} & {\color[HTML]{FF0000} \textbf{0.9958}} & {\color[HTML]{FF0000} \textbf{}} & {\color[HTML]{FF0000} \textbf{45.37}} & {\color[HTML]{FF0000} \textbf{0.9448}} & {\color[HTML]{FF0000} \textbf{}} & {\color[HTML]{FF0000} \textbf{40.73}} & {\color[HTML]{FF0000} \textbf{0.9689}} & {\color[HTML]{FF0000} \textbf{}} & {\color[HTML]{FF0000} \textbf{45.29}} & {\color[HTML]{FF0000} \textbf{0.9698}} \\ \bottomrule
\end{tabular}
}
\end{minipage}
\end{table}

\subsection{All-in-One MedIR Results} 
\label{sec_result_all_in_one}

\textbf{2D All-in-One MedIR.} We evaluate the 2D \textit{All-in-One} setting on the MedIR-2D-500K dataset. We compare it to several general image-restoration methods (SwinIR \cite{liang2021swinir}, Uformer \cite{wang2022uformer}, Restormer \cite{zamir2022restormer}, NAFNet \cite{chen2022nafnet}, Restore-RWKV \cite{yang2025restorerwkv}, and MambaIR \cite{guo2024mambair}) and to SOTA all-in-one approaches (TransWeather \cite{valanarasu2022transweather}, AirNet \cite{li2022airnet}, DRMC \cite{yang2023drmc}, AMIR \cite{yang2024amir}, PromptIR \cite{potlapalli2023promptir}, and AdaIR \cite{cui2025adair}). Tab.~\ref{tab_all_in_one_2D} shows that UniH$^3$ achieves both high efficiency and strong effectiveness. It significantly outperforms all comparison methods across all seven tasks. On average, UniH$^3$ surpasses the second-best AdaIR by 0.21 dB in PSNR, which is an appreciable improvement given that each of the seven task contains thousands of testing images. Visual comparison in Fig.~\ref{fig_all_in_one_2D_vis} demonstrates that UniH$^3$ best restores different types of medical images with higher structural fidelity and finer detail than competing methods. 

\noindent
\textbf{3D All-in-One MedIR.} We assess the UniH$^3$-3D in the 3D \textit{All-in-One} setting on the MedIR-3D-3K dataset. We compare it with several SOTA 3D restoration methods, including 3D-cGAN \cite{wang20183dcgan}, MRDG \cite{wang2020mrdg}, DRMC \cite{yang2023drmc}, Spach Transformer \cite{jang2023spachtransformer}, and Restore-RWKV-3D \cite{yang2025restorerwkv}. Tab.~\ref{tab_all_in_one_3D} shows that UniH$^3$ significantly outperforms all comparison methods across the three 3D MedIR tasks. In particular, UniH$^3$-3D exceeds the second-best Restore-RWKV-3D by an average margin of 0.55 dB in PSNR. Visual comparisons are shown in the \textit{supplement}.

\begin{table}[!t]
\caption{Performance of H$^2$P and H$^2$B on different backbones on the MedIR-2D-500K dataset. $\dag$ denotes applying H$^2$P and H$^2$B to the corresponding backbone.}
\centering
\resizebox{\textwidth}{!}{
\begin{tabular}{ccccccccccccccccccccccccccccc}
\toprule
                                  &  &                                         &  &                                      &  & \multicolumn{2}{c}{\textbf{PET}}                                               & \textbf{} & \multicolumn{2}{c}{\textbf{CT}}                                                & \textbf{} & \multicolumn{2}{c}{\textbf{MRI}}                                               & \textbf{} & \multicolumn{2}{c}{\textbf{X-ray}}                                             & \textbf{} & \multicolumn{2}{c}{\textbf{OCT}}                                               & \textbf{} & \multicolumn{2}{c}{\textbf{Ultrasound}}                                        & \textbf{} & \multicolumn{2}{c}{\textbf{Pathology}}                                         & \textbf{} & \multicolumn{2}{c}{\textbf{Average}}                                           \\ \cline{7-8} \cline{10-11} \cline{13-14} \cline{16-17} \cline{19-20} \cline{22-23} \cline{25-26} \cline{28-29} 
\multirow{-2}{*}{\textbf{Method}} &  & \multirow{-2}{*}{\textbf{\#Params (M)}} &  & \multirow{-2}{*}{\textbf{FLOPs (G)}} &  & \textbf{PSNR↑}                        & \textbf{SSIM↑}                         & \textbf{} & \textbf{PSNR↑}                        & \textbf{SSIM↑}                         & \textbf{} & \textbf{PSNR↑}                        & \textbf{SSIM↑}                         & \textbf{} & \textbf{PSNR↑}                        & \textbf{SSIM↑}                         & \textbf{} & \textbf{PSNR↑}                        & \textbf{SSIM↑}                         & \textbf{} & \textbf{PSNR↑}                        & \textbf{SSIM↑}                         & \textbf{} & \textbf{PSNR↑}                        & \textbf{SSIM↑}                         & \textbf{} & \textbf{PSNR↑}                        & \textbf{SSIM↑}                         \\ \midrule
Uformer                           &  & 50.47                                   &  & 21.42                                &  & 44.51                                 & 0.9874                                 &           & 43.45                                 & 0.9356                                 &           & 39.01                                 & 0.9507                                 &           & 36.59                                 & 0.9353                                 &           & 35.84                                 & 0.8908                                 &           & 27.61                                 & 0.8135                                 &           & 28.55                                 & 0.8011                                 &           & 36.51                                 & 0.9021                                 \\
\rowcolor[HTML]{EFEFEF} 
Uformer $\dag$        &  & 53.15                                   &  & 22.10                                &  & {\color[HTML]{FF0000} \textbf{44.80}} & {\color[HTML]{FF0000} \textbf{0.9880}} &           & {\color[HTML]{FF0000} \textbf{43.59}} & {\color[HTML]{FF0000} \textbf{0.9363}} &           & {\color[HTML]{FF0000} \textbf{39.39}} & {\color[HTML]{FF0000} \textbf{0.9549}} &           & {\color[HTML]{FF0000} \textbf{36.74}} & {\color[HTML]{FF0000} \textbf{0.9354}} &           & {\color[HTML]{FF0000} \textbf{35.94}} & {\color[HTML]{FF0000} \textbf{0.8918}} &           & {\color[HTML]{FF0000} \textbf{27.72}} & {\color[HTML]{FF0000} \textbf{0.8155}} &           & {\color[HTML]{FF0000} \textbf{28.57}} & {\color[HTML]{FF0000} \textbf{0.8022}} &           & {\color[HTML]{FF0000} \textbf{36.68}} & {\color[HTML]{FF0000} \textbf{0.9034}} \\ \hline
Restormer                         &  & 26.12                                   &  & 35.21                                &  & 44.47                                 & 0.9873                                 &           & 43.44                                 & 0.9353                                 &           & 39.05                                 & 0.9515                                 &           & 36.46                                 & 0.9335                                 &           & 35.84                                 & 0.8909                                 &           & 27.66                                 & 0.8138                                 &           & 28.53                                 & 0.8006                                 &           & 36.49                                 & 0.9018                                 \\
\rowcolor[HTML]{EFEFEF} 
Restormer $\dag$      &  & 27.56                                   &  & 35.83                                &  & {\color[HTML]{FF0000} \textbf{44.78}} & {\color[HTML]{FF0000} \textbf{0.9880}} &           & {\color[HTML]{FF0000} \textbf{43.58}} & {\color[HTML]{FF0000} \textbf{0.9362}} &           & {\color[HTML]{FF0000} \textbf{39.37}} & {\color[HTML]{FF0000} \textbf{0.9548}} &           & {\color[HTML]{FF0000} \textbf{36.71}} & {\color[HTML]{FF0000} \textbf{0.9352}} &           & {\color[HTML]{FF0000} \textbf{35.93}} & {\color[HTML]{FF0000} \textbf{0.8919}} &           & {\color[HTML]{FF0000} \textbf{27.73}} & {\color[HTML]{FF0000} \textbf{0.8159}} &           & {\color[HTML]{FF0000} \textbf{28.60}} & {\color[HTML]{FF0000} \textbf{0.8027}} &           & {\color[HTML]{FF0000} \textbf{36.67}} & {\color[HTML]{FF0000} \textbf{0.9035}} \\ \hline
PromptIR                          &  & 35.59                                   &  & 39.49                                &  & 44.52                                 & 0.9874                                 &           & 43.48                                 & 0.9355                                 &           & 39.13                                 & 0.9524                                 &           & 36.57                                 & 0.9341                                 &           & 35.84                                 & 0.8909                                 &           & 27.69                                 & 0.8152                                 &           & 28.54                                 & 0.8006                                 &           & 36.54                                 & 0.9023                                 \\
\rowcolor[HTML]{EFEFEF} 
PromptIR $\dag$       &  & 36.15                                   &  & 40.68                                &  & {\color[HTML]{FF0000} \textbf{44.85}} & {\color[HTML]{FF0000} \textbf{0.9882}} &           & {\color[HTML]{FF0000} \textbf{43.62}} & {\color[HTML]{FF0000} \textbf{0.9366}} &           & {\color[HTML]{FF0000} \textbf{39.40}} & {\color[HTML]{FF0000} \textbf{0.9551}} &           & {\color[HTML]{FF0000} \textbf{36.76}} & {\color[HTML]{FF0000} \textbf{0.9365}} &           & {\color[HTML]{FF0000} \textbf{35.94}} & {\color[HTML]{FF0000} \textbf{0.8919}} &           & {\color[HTML]{FF0000} \textbf{27.75}} & {\color[HTML]{FF0000} \textbf{0.8165}} &           & {\color[HTML]{FF0000} \textbf{28.59}} & {\color[HTML]{FF0000} \textbf{0.8024}} &           & {\color[HTML]{FF0000} \textbf{36.70}} & {\color[HTML]{FF0000} \textbf{0.9039}} \\ \hline
AdaIR                             &  & 28.76                                   &  & 36.74                                &  & 44.55                                 & 0.9875                                 &           & 43.49                                 & 0.9356                                 &           & 39.17                                 & 0.9527                                 &           & 36.60                                 & 0.9344                                 &           & 35.86                                 & 0.8915                                 &           & 27.69                                 & 0.8150                                 &           & 28.56                                 & 0.8015                                 &           & 36.56                                 & 0.9026                                 \\
\rowcolor[HTML]{EFEFEF} 
AdaIR $\dag$          &  & 29.77                                   &  & 37.37                                &  & {\color[HTML]{FF0000} \textbf{44.93}} & {\color[HTML]{FF0000} \textbf{0.9884}} &           & {\color[HTML]{FF0000} \textbf{43.64}} & {\color[HTML]{FF0000} \textbf{0.9367}} &           & {\color[HTML]{FF0000} \textbf{39.53}} & {\color[HTML]{FF0000} \textbf{0.9563}} &           & {\color[HTML]{FF0000} \textbf{36.82}} & {\color[HTML]{FF0000} \textbf{0.9363}} &           & {\color[HTML]{FF0000} \textbf{35.95}} & {\color[HTML]{FF0000} \textbf{0.8920}} &           & {\color[HTML]{FF0000} \textbf{27.76}} & {\color[HTML]{FF0000} \textbf{0.8165}} &           & {\color[HTML]{FF0000} \textbf{28.61}} & {\color[HTML]{FF0000} \textbf{0.8027}} &           & {\color[HTML]{FF0000} \textbf{36.75}} & {\color[HTML]{FF0000} \textbf{0.9041}} \\ \hline
Baseline                          &  & 27.17                                   &  & 25.50                                &  & 44.52                                 & 0.9874                                 &           & 43.47                                 & 0.9355                                 &           & 39.10                                 & 0.9521                                 &           & 36.39                                 & 0.9330                                 &           & 35.88                                 & 0.8914                                 &           & 27.68                                 & 0.8149                                 &           & 28.57                                 & 0.8018                                 &           & 36.52                                 & 0.9023                                 \\
\rowcolor[HTML]{EFEFEF} 
\textbf{UniH$^3$ (Ours)}          &  & 28.96                                   &  & 26.33                                &  & {\color[HTML]{FF0000} \textbf{44.89}} & {\color[HTML]{FF0000} \textbf{0.9883}} &           & {\color[HTML]{FF0000} \textbf{43.65}} & {\color[HTML]{FF0000} \textbf{0.9368}} &           & {\color[HTML]{FF0000} \textbf{39.55}} & {\color[HTML]{FF0000} \textbf{0.9564}} &           & {\color[HTML]{FF0000} \textbf{36.88}} & {\color[HTML]{FF0000} \textbf{0.9368}} &           & {\color[HTML]{FF0000} \textbf{35.96}} & {\color[HTML]{FF0000} \textbf{0.8921}} &           & {\color[HTML]{FF0000} \textbf{27.80}} & {\color[HTML]{FF0000} \textbf{0.8179}} &           & {\color[HTML]{FF0000} \textbf{28.63}} & {\color[HTML]{FF0000} \textbf{0.8035}} &           & {\color[HTML]{FF0000} \textbf{36.77}} & {\color[HTML]{FF0000} \textbf{0.9045}} \\ \bottomrule
\end{tabular}
} 
\label{tab_backbone}
\end{table} 

\begin{table}[!t]
\centering

\begin{minipage}{0.48\textwidth}
\centering
\caption{Component analysis.}
\resizebox{\textwidth}{!}{
\begin{tabular}{clclclclclc}
\toprule
\textbf{H$^2$M} &  & \textbf{H$^2$B} &  & \textbf{\#Params (M)} &  & \textbf{FLOPs (G)} &  & \textbf{PSNR↑} &  & \textbf{SSIM↑} \\ 
\midrule
             &  &              &  & 27.17 &  & 25.50 &  & 36.52 &  & 0.9023 \\
$\checkmark$ &  &              &  & 28.96 &  & 26.33 &  & {\color[HTML]{0000FF}36.66} &  & {\color[HTML]{0000FF}0.9034} \\
             &  & $\checkmark$ &  & 27.17 &  & 25.50 &  & 36.64 &  & 0.9033 \\
\rowcolor[HTML]{EFEFEF}
$\checkmark$ &  & $\checkmark$ &  & 28.96 &  & 26.33 &  & {\color[HTML]{FF0000}\textbf{36.77}} &  & {\color[HTML]{FF0000}\textbf{0.9045}} \\
\bottomrule
\end{tabular}
}
\label{tab_componnet_analysis}
\end{minipage}
\hfill
\begin{minipage}{0.48\textwidth}
\centering
\caption{Ablation studies on HGA.}
\resizebox{\textwidth}{!}{
\begin{tabular}{ccccc}
\toprule
\textbf{Method} & \textbf{\#Params (M)} & \textbf{FLOPs (G)} & \textbf{PSNR↑} & \textbf{SSIM↑} \\
\midrule
w/o HGA         & 27.17 & 25.50 & 36.52 & 0.9023 \\
SFT \cite{wang2018sft}             & 37.46 & 36.99 & {\color[HTML]{0000FF}36.74} & {\color[HTML]{0000FF}0.9041} \\
Cross Attention \cite{cui2025adair} & 29.60 & 27.43 & {\color[HTML]{0000FF}36.67} & {\color[HTML]{0000FF}0.9035} \\
\rowcolor[HTML]{EFEFEF}
\textbf{HGA (Ours)} & 28.96 & 26.33 & {\color[HTML]{FF0000}\textbf{36.77}} & {\color[HTML]{FF0000}\textbf{0.9045}} \\
\bottomrule
\end{tabular}
}
\label{tab_HGA}
\end{minipage}

\end{table}

\begin{table}[!t]
\centering

\begin{minipage}{0.48\textwidth}
\centering
\caption{Ablation studies on H$^2$M.}
\resizebox{\textwidth}{!}{
\begin{tabular}{cccccc}
\toprule
\textbf{\begin{tabular}[c]{@{}c@{}}Inter-task\\ Homogeneity\end{tabular}} & 
\textbf{\begin{tabular}[c]{@{}c@{}}Intra-task\\ Homogeneity\end{tabular}} & 
\textbf{\#Params (M)} & \textbf{FLOPs (G)} & \textbf{PSNR↑} & \textbf{SSIM↑} \\ 
\midrule
 &  & 28.96 & 26.33 & 36.52 & 0.9023 \\
$\checkmark$ &  & 28.96 & 26.33 & 36.61 & 0.9031 \\
 & $\checkmark$ & 28.96 & 26.33 & {\color[HTML]{0000FF} 36.72} & {\color[HTML]{0000FF} 0.9038} \\
\rowcolor[HTML]{EFEFEF}
$\checkmark$ & $\checkmark$ & 28.96 & 26.33 & {\color[HTML]{FF0000} \textbf{36.77}} & {\color[HTML]{FF0000} \textbf{0.9045}} \\
\bottomrule
\end{tabular}
}
\label{tab_H2M}
\end{minipage}
\hfill
\begin{minipage}{0.48\textwidth}
\centering
\caption{Ablation studies on H$^2$B.}
\resizebox{\textwidth}{!}{
\begin{tabular}{cccccc}
\toprule
\textbf{\begin{tabular}[c]{@{}c@{}}Inter-task\\ Heterogeneity\end{tabular}} & 
\textbf{\begin{tabular}[c]{@{}c@{}}Intra-task\\ Heterogeneity\end{tabular}} & 
\textbf{\#Params (M)} & \textbf{FLOPs (G)} & \textbf{PSNR↑} & \textbf{SSIM↑} \\ 
\midrule
 &  & 28.96 & 26.33 & 36.66 & 0.9034 \\
$\checkmark$ &  & 28.96 & 26.33 & {\color[HTML]{0000FF} 36.70} & {\color[HTML]{0000FF} 0.9036} \\
 & $\checkmark$ & 28.96 & 26.33 & {\color[HTML]{0000FF} 36.73} & {\color[HTML]{0000FF} 0.9041} \\
\rowcolor[HTML]{EFEFEF}
\textbf{$\checkmark$} & $\checkmark$ & 28.96 & 26.33 & {\color[HTML]{FF0000} \textbf{36.77}} & {\color[HTML]{FF0000} \textbf{0.9045}} \\
\bottomrule
\end{tabular}
}
\label{tab_H2B}
\end{minipage}

\end{table}

\subsection{Single-Task MedIR Results} 
\label{sec_result_single}

\textbf{2D Single-Task MedIR.} We evaluate UniH$^3$ for 2D single-task MedIR on the MedIR-2D-500K dataset, comparing it to five general image restoration methods. As shown in Tab.~\ref{tab_single_2D}, UniH$^3$ significantly outperforms all comparison methods. On average across seven tasks, UniH$^3$ improves PSNR by 0.15 dB over the second-best MambaIR.

\noindent
\textbf{3D Single-Task MedIR.} We evaluate 3D single-task MedIR on the MedIR-3D-3K dataset and compare UniH$^3$-3D to five SOTA 3D restoration methods. As shown in Tab.~\ref{tab_single_3D}, UniH$^3$-3D beats all comparison methods. In particular, it improves average PSNR by 1.48 dB over the second-best Spach Transformer~\cite{jang2023spachtransformer} across the three tasks.

\subsection{Ablation Studies} 
\label{sec_ablation}
To evaluate the effectiveness of individual components, we conduct ablation experiments on the 2D \textit{All-in-One} MedIR task with the MedIR-2D-500K dataset.

\begin{figure}[!t]
\centering
\begin{minipage}[c]{0.4\textwidth}
    \centering
    \includegraphics[width=\textwidth]{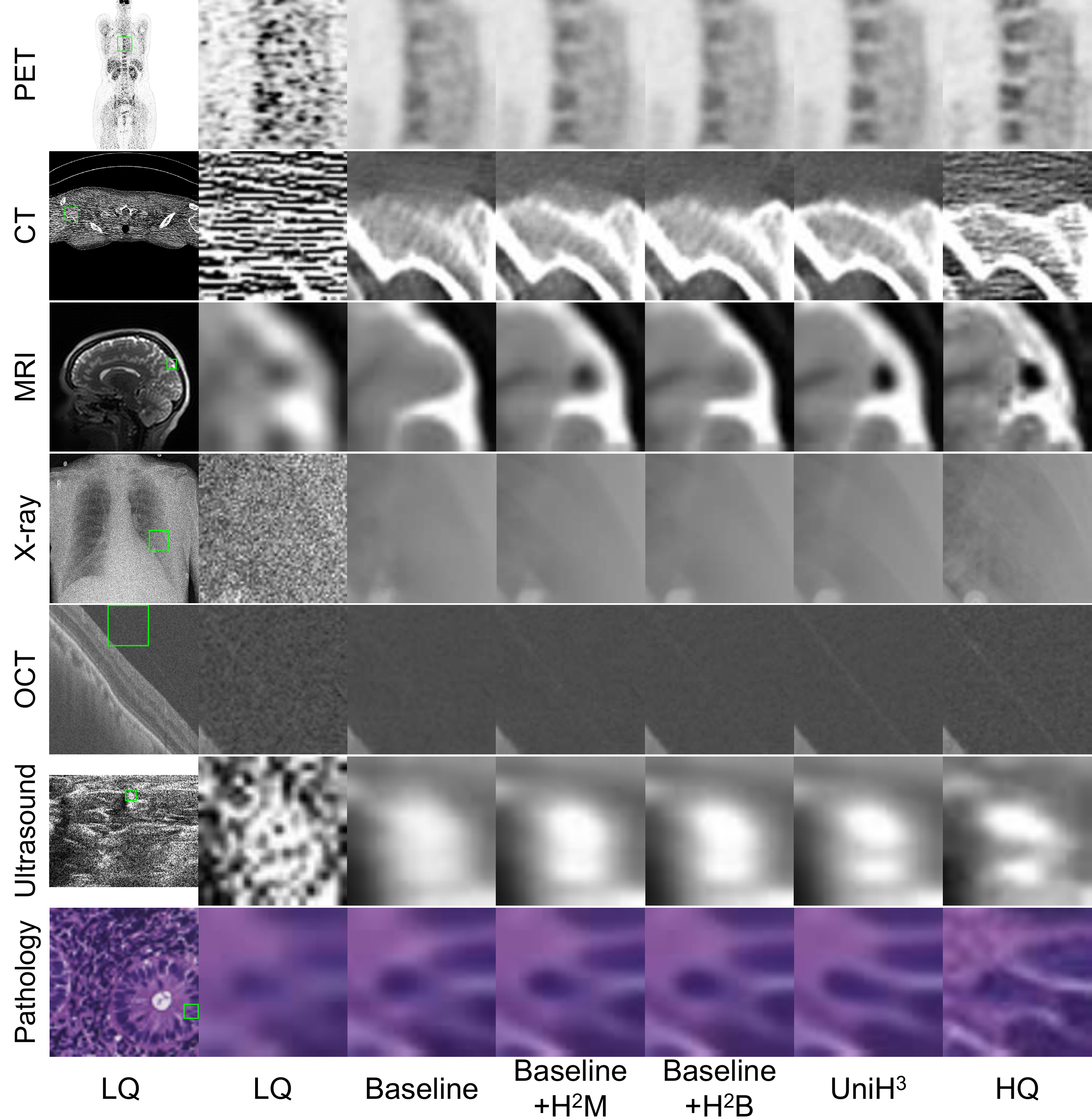}
    \caption{Visual comparison for component analysis.}
    \label{fig_ablation_study}
\end{minipage}
\hfill
\begin{minipage}[c]{0.55\textwidth}
    \centering
    \includegraphics[width=\textwidth]{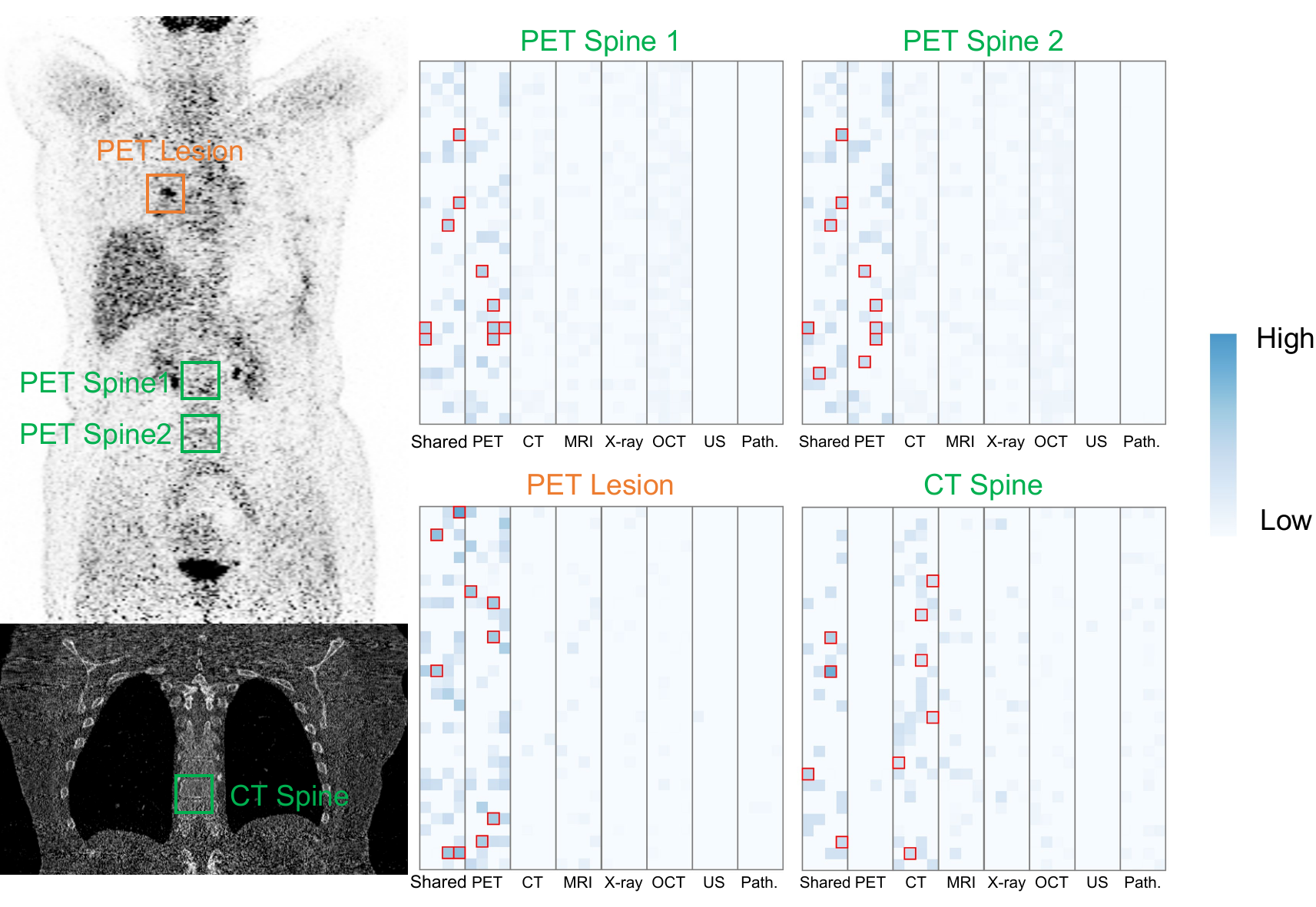}
    \caption{Retrieval attention map in H$^2$M. The Top-10 scores are marked by red rectangles.}
    \label{fig_attentionmap}
\end{minipage}
\end{figure}

\noindent
\textbf{Component Analysis of H$^2$M and H$^2$B.} We first perform a component analysis of the H$^2$M module and the H$^2$B strategy by selectively disabling each component. To disable H$^2$M, we remove the H$^2$M module and replace the HGA mechanism with a transposed self-attention layer \cite{zamir2022restormer}. The H$^2$B is disabled by substituting the loss term $\mathcal{L}_{\text{H$^2$B}}$ with an $L_1$ loss. Table~\ref{tab_componnet_analysis} shows that both components improve model performance with minimal increase in computational cost. This finding is further supported by the visual comparison in Fig.~\ref{fig_ablation_study}, where both components contribute to better preservation of image details. Moreover, we apply the H$^2$M module and H$^2$B strategy to other Transformer-based U-shaped backbones, including Uformer, Restormer, PromptIR, and AdaIR. Results in the Tab.~\ref{tab_backbone} demonstrate significant improvements across all these backbones.

\noindent
\textbf{Ablation Studies on H$^2$M.} We investigate the effectiveness of both inter- and intra-task homogeneity priors in H$^2$M by selectively disabling the task-specific and task-shared slots in $M$. By replacing these slots with naive learnable parameters—thus isolating them from the HQ Homogeneity Distillation procedure—we observe a drop in performance in Tab.~\ref{tab_H2M}. Results indicate that both priors independently benefit the restoration process, and their combination achieves the best performance. To further understand the internal mechanics of H$^2$M, Fig.~\ref{fig_attentionmap} visualizes the retrieval attention maps for PET and CT tokens. The distributions confirm that tokens primarily query the task-shared slot and their specific modality slot. Furthermore, attention maps reflect clear semantic correlations: anatomically similar tokens within the same modality share highly similar query patterns (8/10 top-score overlap for two PET spine tokens), and cross-modality similarities are also captured (2/10 overlap for PET and CT spine tokens). In contrast, dissimilar anatomies exhibit distinct query patterns (0/10 overlap for PET spine and lesion tokens). This provides strong visual evidence that H$^2$M successfully maps, stores, and retrieves hierarchical anatomical priors.

\noindent
\textbf{Ablation Studies on HGA.} We validate the impact of HGA by replacing it with alternative guiding mechanisms, including Spatial Feature Transform (SFT) \cite{wang2018sft} and cross attention \cite{cui2025adair}. Tab.~\ref{tab_HGA} shows that the proposed HGA performs the best with minimal computation and parameter increase. 

\begin{figure}[!t]
\centering
\begin{minipage}[t]{0.5\textwidth}
    \centering
    \includegraphics[width=\linewidth]{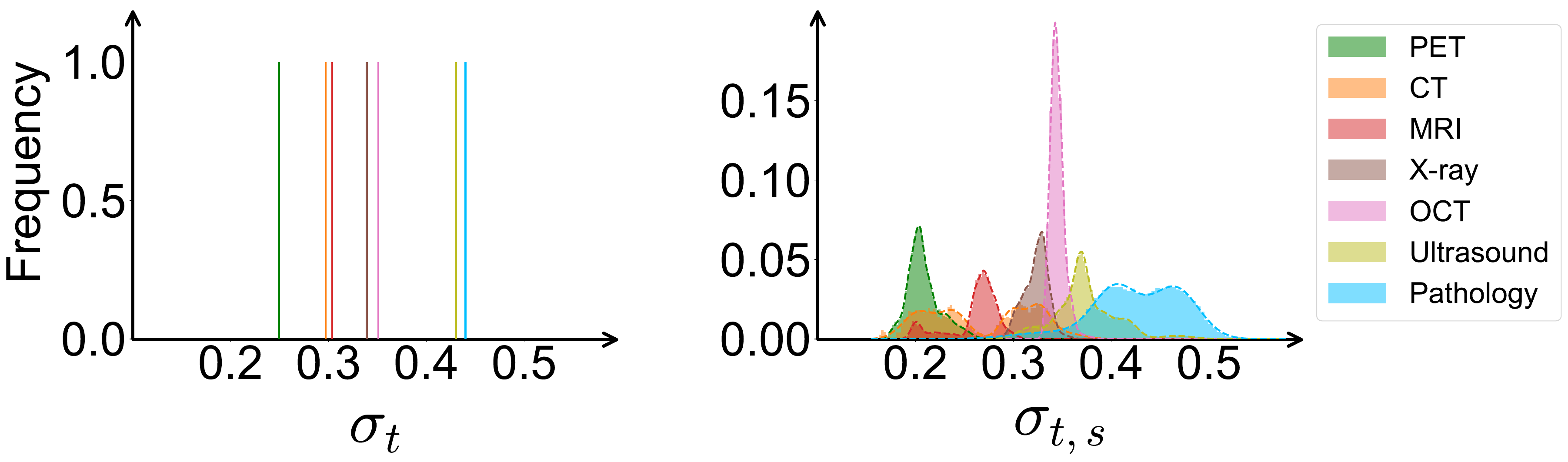}
    \caption{Estimated uncertainty distribution.}
    \label{fig_uncertainty}
\end{minipage}
\hfill
\begin{minipage}[t]{0.49\textwidth}
    \centering
    \includegraphics[width=\linewidth]{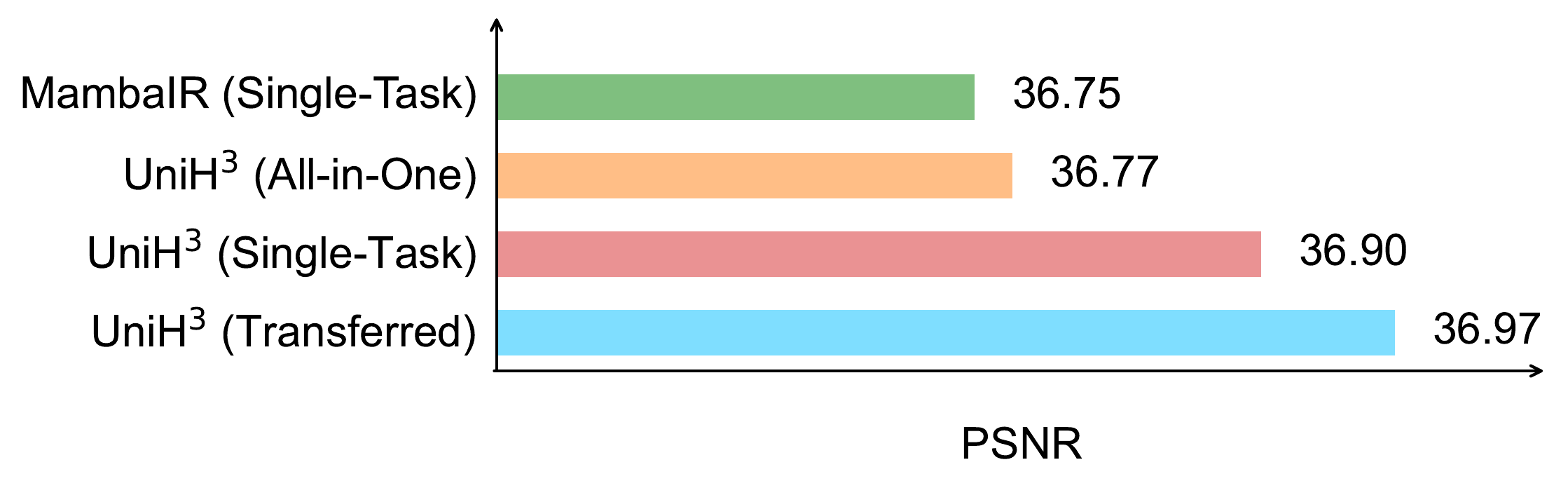}
    \caption{All-in-One vs. Single-Task.}
    \label{fig_vs_hist}
\end{minipage}
\end{figure}

\noindent
\textbf{Ablation Studies on H$^2$B.} We validate the roles of intra- and inter-task heterogeneity in H$^2$B in Tab.~\ref{tab_H2B}. The results indicate that mitigating both levels of heterogeneity improves overall restoration performance, with their joint optimization achieving the best results. In Fig.~\ref{fig_uncertainty}, we visualize the estimated uncertainty distributions. While the conventional $\mathcal{L}_{\text{UB}}$ \cite{kendall2018uncertainty_loss} estimates an uncertainty $\sigma_t$ per task and therefore handles only inter-task heterogeneity, our proposed $\mathcal{L}_{\text{H}^2\text{B}}$ estimates uncertainty $\sigma_{t,s}$ at both the task and the sample level: each sample receives its own uncertainty while each task exhibits a distinct uncertainty distribution. This enables our $\mathcal{L}_{\text{H}^2\text{B}}$ to capture finer-grained relationships, i.e., both inter-task and intra-task heterogeneity, facilitating convergence toward a more optimal solution for diverse MedIR tasks.

\section{Discussion and Limitation}
All-in-one medical image restoration is an emerging research field. The practical value of all-in-one models remains under active discussion. In this paper, experiments on a large-scale dataset show that a single all-in-one model, UniH$^3$, already achieves comparable performance to SOTA single-task MambaIR models across seven MedIR tasks (see Fig.~\ref{fig_vs_hist}). This result strongly supports the practicality of developing all-in-one models instead of separate single-task models for MedIR tasks. Additionally, by fine-tuning the well-trained all-in-one UniH$^3$ model for each task, we are able to transfer its learned knowledge to individual tasks and achieve further improvements (also shown in Fig.~\ref{fig_vs_hist}), indicating that the all-in-one model can serve as a transferable pretrained backbone. Our study has limitations: we focus only on the primary restoration task within each modality and therefore do not cover other tasks or degradation types that may occur in the same modality. Future work should address these gaps and pursue more universal MedIR models to benefit clinical diagnosis and more downstream tasks~\cite{zhou2024sdpt,zhou2026sdpt,wu2024attriprompter,wu2025visual,hao2026styledrive,chen2026beyond,zhou2026comprehensive,zhou25phi,zhou2023video,zhou2025asal}.

%



\section{Conclusion}
In this paper, we have presented UniH$^3$, a novel and unified framework for all-in-one medical image restoration. By moving beyond the conventional focus on inter-task heterogeneity, UniH$^3$ effectively leverages the inherent hierarchical homogeneity across diverse medical imaging modalities. The integration of the Hierarchical Homogeneity Memory (H$^2$M) module and the Homogeneity-Guided Attention (HGA) mechanism allows the model to distill and utilize inter- and intra-task homogeneity priors to guide the restoration process. Simultaneously, the Hierarchical Heterogeneity Balancer (H$^2$B) ensures a stable and balanced multi-task learning process by mitigating both inter- and intra-task heterogeneity. Extensive evaluations on the large-scale MedIR-2D-500K and MedIR-3D-3K benchmarks demonstrate that UniH$^3$ significantly outperforms existing methods, achieving state-of-the-art performance in both all-in-one and single-task scenarios. In future work, we plan to expand the task spectrum by incorporating additional modalities and degradation types, moving toward more general MedIR models.

\section*{Acknowledgements}
This work is supported by the National Natural Science Foundation in China under Grant U23B2063 and 62371016, the Bejing Natural Science Foundation Haidian District Joint Fund in China under Grant L2602042, the Beijing hope run special fund of cancer foundation of China under Grant LC2018L02, the Fundamental Research Funds for the Central University of China from the State Key Laboratory of Software Development Environment in Beihang University in China, the 111 Proiect in China under Grant B13003, the Academic Excellence Foundation of BUAA for PhD Students.

\bibliographystyle{splncs04}
\bibliography{main}

\newpage
\appendix  

\renewcommand{\thetable}{\Roman{table}}
\renewcommand{\thefigure}{\Roman{figure}}

\setcounter{table}{0}
\setcounter{figure}{0}


\section{Availability of Code and Data}
The code and data are released at \href{https://github.com/Yaziwel/UniH3.git}{https://github.com/Yaziwel/UniH3}. We hope this study will contribute to the advancement of general-purpose MedIR methods.

\section{Transposed Self-Attention-Based HGA}
Let the query, key, and value be $Q,K,V\in\mathbb{R}^{H'W'\times C'}$. Following Eqs.\ref{eq_attn}-\ref{eq_HGA} in Sec.~\ref{sec_HGA}, we derive the transposed self-attention–based \cite{zamir2022restormer} homogeneity-guided attention (HGA) as follows:
\begin{equation}
H_0=VA, \qquad A=\operatorname{Softmax}(\frac{Q^{\mathsf{T}}K}{\sqrt{C'}}).
\end{equation} 
\begin{equation}
\begin{aligned}
H_1=(V+V^{H})A=VA+V^{H}A.
\end{aligned}
\label{eq_residual_value2}
\end{equation} 
\begin{equation}
\begin{aligned}
H_2&=V(A-I)+V^{H}(A+I)\\
&=(V+V^{H})A+V^{H}-V.
\end{aligned}
\label{eq_HGA_init2}
\end{equation} 
\begin{equation}
H=[(1-\lambda_1)V+\lambda_1V^{H}]A+\lambda_2(V^{H}-V).
\label{eq_HGA2}
\end{equation} 
The finally derived transposed self-attention-based HGA in Eq.~\ref{eq_HGA2} can also be illustrated by Fig.~\ref{fig_hga}, which augments attention with simple addition and subtraction operations on the value. Therefore, Fig.~\ref{fig_hga} illustrates both the self-attention–based HGA and the transposed self-attention–based HGA. 

\begin{table}[!t]
\caption{MedIR-2D-500K and MedIR-3D-3K datasets.}
\centering
\resizebox{\textwidth}{!}{
\begin{tabular}{ccccccc}
\toprule
\textbf{Dataset}                          & \textbf{Dimension}    & \textbf{Modality}                                 & \textbf{Training}                        & \textbf{Testing}                        & \textbf{Total}                           & \textbf{Data Source}                                           \\ \midrule
                                          &                       &                                                   & 73,125                                   & 8,000                                   & 81,125                                   & \cite{xue2022cross}                           \\
                                          &                       &                                                   & 1,850                                    & 300                                     & 2,150                                    & Private1                                                       \\
                                          &                       & \multirow{-3}{*}{PET}                             & 2,025                                    & 300                                     & 2,325                                    & Private2                                                       \\
                                          &                       & \cellcolor[HTML]{EFEFEF}\textbf{PET Total}        & \cellcolor[HTML]{EFEFEF}\textbf{77,000}  & \cellcolor[HTML]{EFEFEF}\textbf{8,600}  & \cellcolor[HTML]{EFEFEF}\textbf{85,600}  & \cellcolor[HTML]{EFEFEF}-                                      \\
                                          &                       &                                                   & 4,470                                    & 1,466                                   & 5,936                                    & \cite{mccollough2017aapm}                     \\
                                          &                       &                                                   & 22,995                                   & 2,534                                   & 25,529                                   & \cite{moen2021ldct}                           \\
                                          &                       & \multirow{-3}{*}{CT}                              & 36,535                                   & 4,100                                   & 40,635                                   & Private3                                                       \\
                                          &                       & \cellcolor[HTML]{EFEFEF}\textbf{CT Total}         & \cellcolor[HTML]{EFEFEF}\textbf{64,000}  & \cellcolor[HTML]{EFEFEF}\textbf{8,100}  & \cellcolor[HTML]{EFEFEF}\textbf{72,100}  & \cellcolor[HTML]{EFEFEF}-                                      \\
                                          &                       &                                                   & 59,600                                   & 6,660                                   & 66,260                                   & \cite{van2013HCP}                             \\
                                          &                       & \multirow{-2}{*}{MRI}                             & 15,400                                   & 1,740                                   & 17,140                                   & \cite{ixi_dataset}                           \\
                                          &                       & \cellcolor[HTML]{EFEFEF}\textbf{MRI Total}        & \cellcolor[HTML]{EFEFEF}\textbf{75,000}  & \cellcolor[HTML]{EFEFEF}\textbf{8,400}  & \cellcolor[HTML]{EFEFEF}\textbf{83,400}  & \cellcolor[HTML]{EFEFEF}-                                      \\
                                          &                       &                                                   & 100,600                                  & 11,000                                  & 111,600                                  & \cite{wang2017chestxray}                      \\
                                          &                       &                                                   & 3,000                                    & 200                                     & 3,200                                    & \cite{chowdhury2020covid,rahman2021exploring} \\
                                          &                       & \multirow{-3}{*}{X-ray}                           & 4,400                                    & 400                                     & 4,800                                    & \cite{wang2023medfmc}                         \\
                                          &                       & \cellcolor[HTML]{EFEFEF}\textbf{X-ray Total}      & \cellcolor[HTML]{EFEFEF}\textbf{108,000} & \cellcolor[HTML]{EFEFEF}\textbf{11,600} & \cellcolor[HTML]{EFEFEF}\textbf{119,600} & \cellcolor[HTML]{EFEFEF}-                                      \\
                                          &                       &                                                   & 31932                                    & 3592                                    & 35,524                                   & \cite{li2024octa}                             \\
                                          &                       &                                                   & 33                                       & 4                                       & 37                                       & \cite{geng2022pku}                            \\
                                          &                       & \multirow{-3}{*}{OCT}                             & 35                                       & 4                                       & 39                                       & \cite{fang2012sbsdi}                          \\
                                          &                       & \cellcolor[HTML]{EFEFEF}\textbf{OCT Total}        & \cellcolor[HTML]{EFEFEF}\textbf{32000}   & \cellcolor[HTML]{EFEFEF}\textbf{3600}   & \cellcolor[HTML]{EFEFEF}\textbf{35,600}  & \cellcolor[HTML]{EFEFEF}-                                      \\
                                          &                       &                                                   & 1,950                                    & 490                                     & 2,440                                    & \cite{yiguo2023usenhance}                     \\
                                          &                       &                                                   & 8,900                                    & 2,220                                   & 11,120                                   & \cite{ultrasound-nerve-segmentation}          \\
                                          &                       &                                                   & 1,050                                    & 260                                     & 1,310                                    & \cite{van2018hc}                              \\
                                          &                       &                                                   & 13,600                                   & 1,680                                   & 15,280                                   & \cite{leclerc2019camus}                       \\
                                          &                       & \multirow{-5}{*}{Ultrasound}                      & 30,500                                   & 1,150                                   & 31,650                                   & \cite{yang2023cardiac}                        \\
                                          &                       & \cellcolor[HTML]{EFEFEF}\textbf{Ultrasound Total} & \cellcolor[HTML]{EFEFEF}\textbf{56,000}  & \cellcolor[HTML]{EFEFEF}\textbf{5,800}  & \cellcolor[HTML]{EFEFEF}\textbf{61,800}  & \cellcolor[HTML]{EFEFEF}-                                      \\
                                          &                       &                                                   & 13200                                    & 1,830                                   & 15,030                                   & \cite{drifka2016tma}                          \\
                                          &                       &                                                   & 350                                      & 90                                      & 440                                      & \cite{kumar2019monuseg}                       \\
                                          &                       &                                                   & 23100                                    & 2,300                                   & 25,400                                   & \cite{da2022digestpath}                       \\
                                          &                       &                                                   & 400                                      & 35                                      & 435                                      & \cite{sirinukunwattana2017glas}               \\
                                          &                       &                                                   & 7650                                     & 695                                     & 8,345                                    & \cite{aksac2019brecahad}                      \\
                                          &                       & \multirow{-6}{*}{Pathology}                       & 1,300                                    & 150                                     & 1,450                                    & \cite{tekin2023tubule}                        \\
                                          &                       & \cellcolor[HTML]{EFEFEF}\textbf{Pathology Total}  & \cellcolor[HTML]{EFEFEF}\textbf{46000}   & \cellcolor[HTML]{EFEFEF}\textbf{5100}   & \cellcolor[HTML]{EFEFEF}\textbf{51,100}  & \cellcolor[HTML]{EFEFEF}-                                      \\
\multirow{-33}{*}{\textbf{MedIR-2D-500K}} & \multirow{-33}{*}{2D} & \cellcolor[HTML]{EFEFEF}\textbf{Total}            & \cellcolor[HTML]{EFEFEF}\textbf{458,000} & \cellcolor[HTML]{EFEFEF}\textbf{51,200} & \cellcolor[HTML]{EFEFEF}\textbf{509,200} & \cellcolor[HTML]{EFEFEF}                                       \\ \hline
                                          &                       &                                                   & 1,233                                    & 138                                     & 1,371                                    & \cite{xue2022cross}                           \\
                                          &                       &                                                   & 74                                       & 9                                       & 83                                       & Private1                                                       \\
                                          &                       & \multirow{-3}{*}{PET}                             & 81                                       & 9                                       & 90                                       & Private2                                                       \\
                                          &                       & \cellcolor[HTML]{EFEFEF}\textbf{PET Total}        & \cellcolor[HTML]{EFEFEF}\textbf{1,388}   & \cellcolor[HTML]{EFEFEF}\textbf{156}    & \cellcolor[HTML]{EFEFEF}\textbf{1,544}   & \cellcolor[HTML]{EFEFEF}-                                      \\
                                          &                       &                                                   & 8                                        & 2                                       & 10                                       & \cite{mccollough2017aapm}                     \\
                                          &                       &                                                   & 135                                      & 15                                      & 150                                      & \cite{moen2021ldct}                           \\
                                          &                       & \multirow{-3}{*}{CT}                              & 115                                      & 13                                      & 128                                      & Private3                                                       \\
                                          &                       & \cellcolor[HTML]{EFEFEF}\textbf{CT Total}         & \cellcolor[HTML]{EFEFEF}\textbf{258}     & \cellcolor[HTML]{EFEFEF}\textbf{30}     & \cellcolor[HTML]{EFEFEF}\textbf{288}     & \cellcolor[HTML]{EFEFEF}-                                      \\
                                          &                       &                                                   & 519                                      & 58                                      & 577                                      & \cite{van2013HCP}                             \\
                                          &                       & \multirow{-2}{*}{MRI}                             & 1001                                     & 112                                     & 1,113                                    & \cite{ixi_dataset}                           \\
                                          &                       & \cellcolor[HTML]{EFEFEF}\textbf{MRI Total}        & \cellcolor[HTML]{EFEFEF}\textbf{1,520}   & \cellcolor[HTML]{EFEFEF}\textbf{170}    & \cellcolor[HTML]{EFEFEF}\textbf{1,690}   & \cellcolor[HTML]{EFEFEF}-                                      \\
\multirow{-12}{*}{\textbf{MedIR-3D-3K}}   & \multirow{-12}{*}{3D} & \cellcolor[HTML]{EFEFEF}\textbf{Total}            & \cellcolor[HTML]{EFEFEF}\textbf{3,166}   & \cellcolor[HTML]{EFEFEF}\textbf{356}    & \cellcolor[HTML]{EFEFEF}\textbf{3,522}   & \cellcolor[HTML]{EFEFEF}-                                      \\ \bottomrule
\end{tabular}
} 
\label{tab_dataset_2}
\end{table} 

\section{Additional Dataset Information} 
The detailed dataset information is shown in Tab.~\ref{tab_dataset_2}. We then describe the information of private data and the methods used to simulate LQ–HQ image pairs.
\subsection{Private Data Source} 
We collected two private datasets (Private1 and Private2 in Tab.~\ref{tab_dataset_2}) for PET, and one private dataset (Private3 in Tab.~\ref{tab_dataset_2}) for CT. This study and the experimental procedures involving all three private datasets were approved by the Biological and Medical Ethnics Committee of Beihang University (approval number BM20250008). Informed consent was obtained from all participating patients. 

\noindent
\textbf{Private1.} We collect 83 3D whole-body PET images using PolarStar m660 PET imaging system, whith an average administered dose of 293 MBq of $^{18}$F-FDG. 2D images are extracted from slices of 3D images, excluding slices without anatomical content (e.g., air-only regions).

\noindent
\textbf{Private2.} We collect 90 3D whole-body PET images using PolarStar Flight PET imaging system, whith an average administered dose of 301 MBq of $^{18}$F-FDG. 2D images are extracted from slices of 3D images, excluding slices without anatomical content (e.g., air-only regions). 

\noindent
\textbf{Private3.} We collect 128 3D CT images using the Sinovision CT imaging system. Among them, 18 images are of the spine, 50 of the lungs, and 60 of soft tissues. 2D images are extracted from slices of 3D images, excluding slices without anatomical content (e.g., air-only regions).

\subsection{Methods for Generating LQ-HQ image Pairs}
We describe the simulation methods used to generate LQ–HQ image pairs for different tasks in Tab.~\ref{tab_dataset_2}. Our focus is primarily on the key degradation affecting each imaging modality.

\noindent
\textbf{PET image denoising.} Following the paper \cite{xue2022cross}, the original PET data is collected in listmode. To simulate LQ PET images, list mode data are randomly subsampled to achieve a dose reduction factor of 10. Both HQ and LQ PET images undergo reconstruction using the standard OSEM method. 

\noindent
\textbf{CT Image Denoising.} Following the paper \cite{mccollough2017aapm}, we first obtain the original CT projection data. To simulate LQ CT images, Poisson noise is inserted into the projection data for each case to reach a noise level that corresponded to 25\% of the full dose. Both HQ and LQ CT images are then generated using standard CT reconstruction applied to their respective projection data. 

\noindent
\textbf{MRI Image Super-Resolution.} Following the paper \cite{yang2024amir}, the LQ image is generated by transforming the HQ image to the frequency domain, retaining only the central 6.25$\%$ of frequency data points while zero-filling the high-frequency parts, and then converting it back to the image domain. 

\noindent
\textbf{X-ray Image Denoising.} According to previous studies \cite{thanh2019x_ray_denoising}, X-ray images are primarily affected by Poisson noise. Accordingly, we generate LQ images using the following formulation: $I^{LQ}=\frac{\operatorname{Poisson}(\lambda I^{HQ})}{\lambda}$, where we set the noise level to $\lambda=30$. 

\noindent
\textbf{OCT Image Denosing.} According to the paper \cite{dong2020oct_denoising}, OCT images suffer from speckle noise, which can be reduced by averaging repeated scans acquired at the same location. Following prior work \cite{geng2022pku}, we averaged five repeated scans to produce a noise-reduced, HQ image, and randomly selected one of the five original scans as the LQ image. 

\noindent
\textbf{Ultrasound Image Denoising.} Ulrasound image often suffer from a multiplicative speckle noise, which can be approximated as: $I^{LQ}=I^{HQ}+(I^{HQ})^{\gamma}\epsilon$, where $\epsilon$ is zero-mean Gaussian noise and $\gamma$ controls the strength of the content-dependent perturbation. We set $\gamma=0.5$

\noindent
\textbf{Pathological Image Super-Resolution.} We perform 4× Y-channel super-resolution. To synthesize LQ pathological images, we first convert the RGB pathlogical images to the YCbCr color space and then downsample all three channels by a factor of four using bicubic interpolation. During reconstruction, only the Y (luminance) channel is processed by the model, while the Cb and Cr channels are restored using bicubic upsampling, since the human visual system is far more sensitive to luminance than to chrominance.

\section{Additional Experiments.} 
\textbf{2D All-in-One MedIR Results on AMIR Dataset \cite{yang2024amir}.} We also conduct an all-in-one MedIR performance comparison on the dataset provided in the AMIR paper \cite{yang2024amir}. This dataset includes three tasks: PET image denoising, CT image denoising, and MRI image super-resolution. The results are shown in Tab. \ref{tab_all_in_one_AMIR}. Our proposed UniH$^3$ consistently outperforms all comparison methods across these three tasks.

\begin{table}[h]
\caption{2D All-in-One MedIR results on the AMIR datatset \cite{yang2024amir}.}
\centering
\resizebox{\textwidth}{!}{
\begin{tabular}{ccccccccccccc}
\toprule
                                  & \textbf{} & \multicolumn{2}{c}{\textbf{PET}}                                               & \textbf{}                        & \multicolumn{2}{c}{\textbf{CT}}                                                & \textbf{}                        & \multicolumn{2}{c}{\textbf{MRI}}                                               & \textbf{}                        & \multicolumn{2}{c}{\textbf{Average}}                                           \\ \cline{3-4} \cline{6-7} \cline{9-10} \cline{12-13} 
\multirow{-2}{*}{\textbf{Method}} & \textbf{} & \textbf{PSNR↑}                        & \textbf{SSIM↑}                         & \textbf{}                        & \textbf{PSNR↑}                        & \textbf{SSIM↑}                         & \textbf{}                        & \textbf{PSNR↑}                        & \textbf{SSIM↑}                         & \textbf{}                        & \textbf{PSNR↑}                        & \textbf{SSIM↑}                         \\ \midrule
Restormer                         &           & 37.14                                 & 0.9473                                 &                                  & 33.61                                 & 0.9177                                 &                                  & 31.72                                 & 0.9362                                 &                                  & 34.16                                 & 0.9337                                 \\
Eformer                           &           & 35.11                                 & 0.9091                                 &                                  & 32.44                                 & 0.9078                                 &                                  & 29.19                                 & 0.8728                                 &                                  & 32.25                                 & 0.8966                                 \\
Spach Transformer                 &           & 37.05                                 & 0.9445                                 &                                  & 33.47                                 & 0.9155                                 &                                  & 31.18                                 & 0.9290                                 &                                  & 33.90                                 & 0.9297                                 \\
DRMC                              &           & 36.19                                 & 0.9376                                 &                                  & 33.28                                 & 0.9153                                 &                                  & 29.55                                 & 0.9032                                 &                                  & 33.01                                 & 0.9187                                 \\
AirNet                            &           & 37.17                                 & 0.9451                                 & {\color[HTML]{0000FF}}          & 33.62                                 & 0.9176                                 & {\color[HTML]{0000FF}}          & 31.39                                 & 0.9316                                 & {\color[HTML]{0000FF}}          & 34.06                                 & 0.9314                                 \\
AMIR                              &           & {\color[HTML]{0000FF} 37.12}          & {\color[HTML]{0000FF} 0.9475}          & {\color[HTML]{0000FF} \textbf{}} & {\color[HTML]{0000FF} 33.70}          & {\color[HTML]{0000FF} 0.9182}          & {\color[HTML]{0000FF} \textbf{}} & {\color[HTML]{0000FF} 32.03}          & {\color[HTML]{0000FF} 0.9396}          & {\color[HTML]{0000FF} \textbf{}} & {\color[HTML]{0000FF} 34.28}          & {\color[HTML]{0000FF} 0.9351}          \\ \rowcolor[HTML]{EFEFEF} 
\textbf{UniH$^3$ (Ours)}            &           & {\color[HTML]{FF0000} \textbf{37.40}} & {\color[HTML]{FF0000} \textbf{0.9478}} & {\color[HTML]{FF0000} \textbf{}} & {\color[HTML]{FF0000} \textbf{33.79}} & {\color[HTML]{FF0000} \textbf{0.9203}} & {\color[HTML]{FF0000} \textbf{}} & {\color[HTML]{FF0000} \textbf{32.14}} & {\color[HTML]{FF0000} \textbf{0.9405}} & {\color[HTML]{FF0000} \textbf{}} & {\color[HTML]{FF0000} \textbf{34.44}} & {\color[HTML]{FF0000} \textbf{0.9362}} \\ \bottomrule
\end{tabular}
} 
\label{tab_all_in_one_AMIR}
\end{table}

\section{Additional Visualization.}
\begin{figure*}[!t]
\centering
\includegraphics[width=0.95\textwidth]{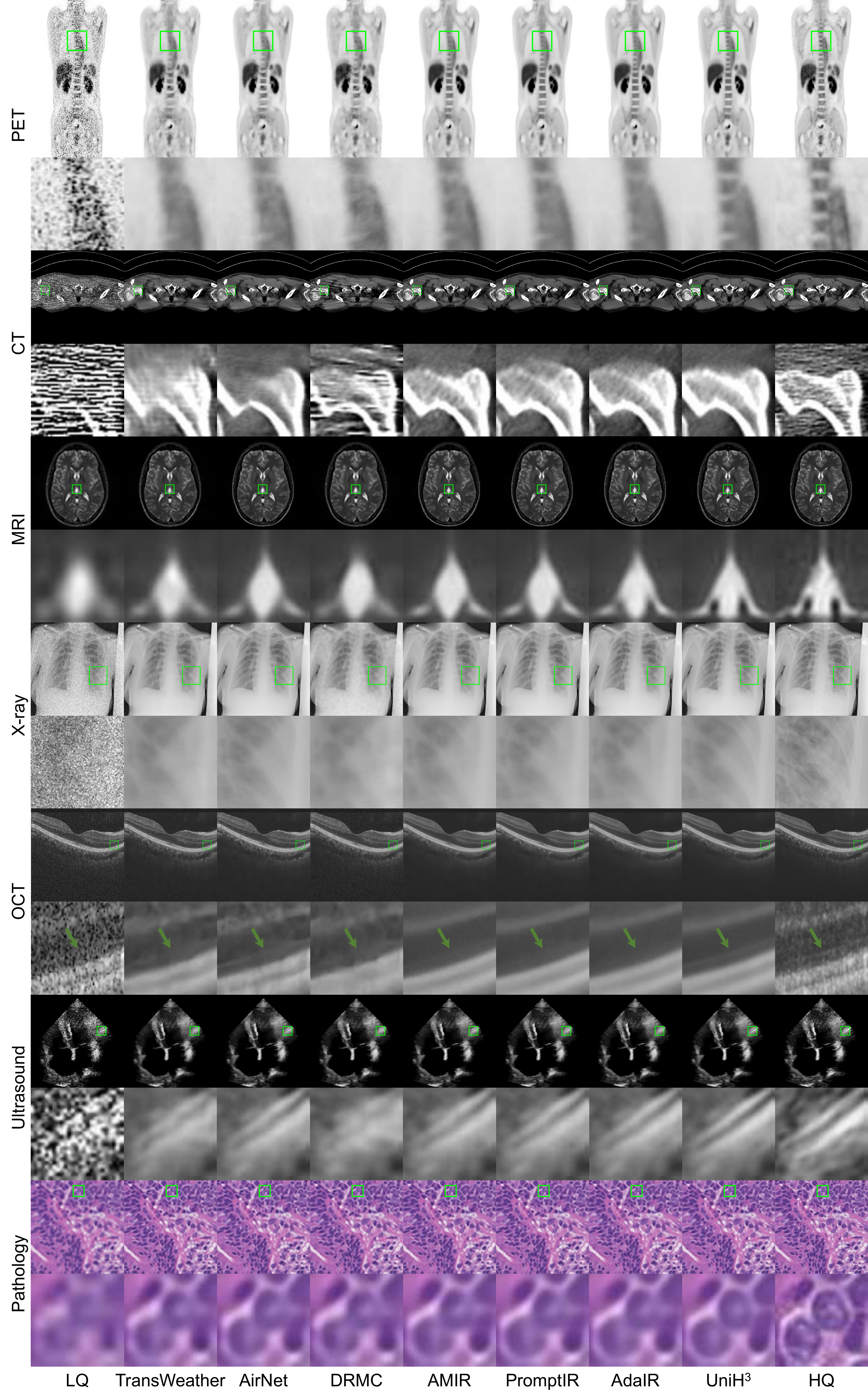}
\caption{Visual comparison for 2D all-in-one MedIR on the MedIR-2D-500K dataset.
} 
\label{vis_comparison_1}
\end{figure*}

\begin{figure*}[!t]
\centering
\includegraphics[width=0.95\textwidth]{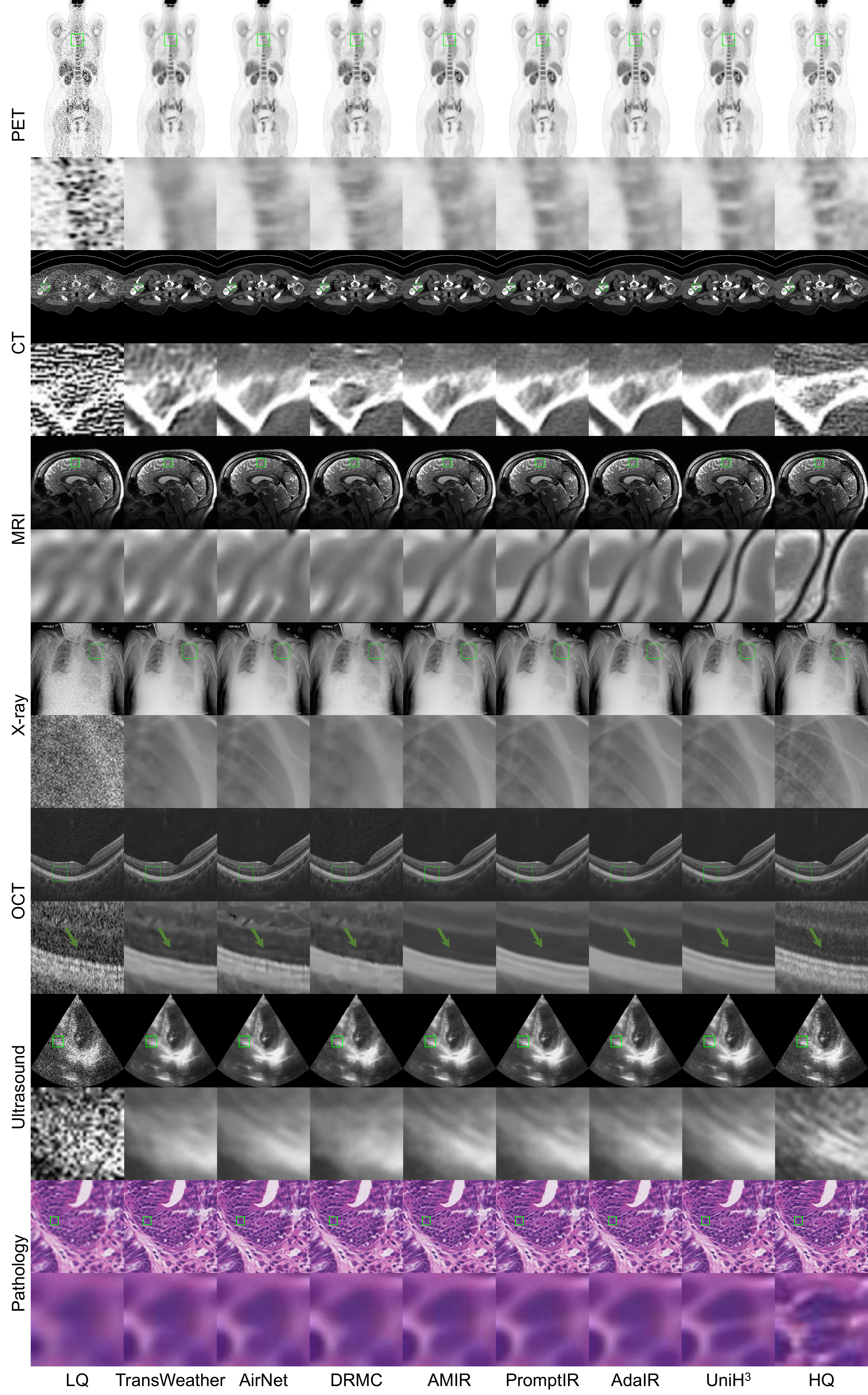}
\caption{Visual comparison for 2D all-in-one MedIR on the MedIR-2D-500K dataset.
} 
\label{vis_comparison_2}
\end{figure*}

\textbf{Visual Comparison for 2D All-in-One MedIR.} We provide an additional visual comparison for 2D all-in-one medical image restoration in Fig.~\ref{vis_comparison_1} and Fig.~\ref{vis_comparison_2}. Our UniH$^3$ best preserves structures and details across seven tasks.

\begin{figure*}[t]
\centering
\includegraphics[width=0.95\textwidth]{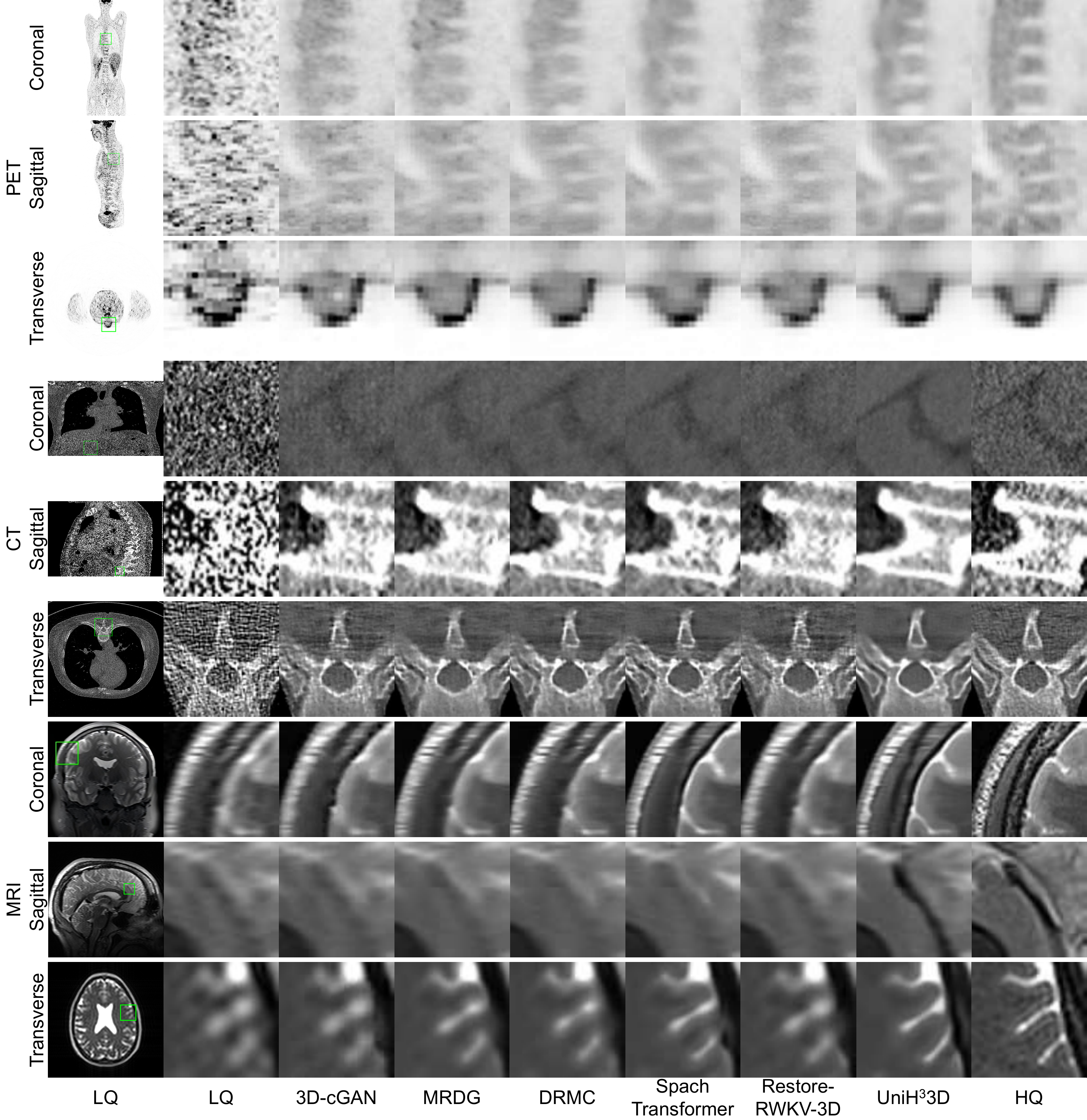}
\caption{Visual comparison for 3D all-in-one MedIR on the MedIR-3D-3K dataset.
} 

\label{fig_all_in_one_3D_vis}
\end{figure*}

\noindent
\textbf{Visual Comparison for 3D All-in-One MedIR.} We provide an additional visual comparison across the coronal, sagittal, and transverse planes for the 3D all-in-one medical image restoration in Fig.~\ref{fig_all_in_one_3D_vis}. Our UniH$^3$-3D best preserves structures and details across three tasks.

\end{document}